%% file: main.tex
\documentclass[10pt,twocolumn,letterpaper]{article}

\usepackage{booktabs}
\usepackage{makecell}
\usepackage{multirow}
\usepackage{graphicx}
\usepackage{tabularx}
\usepackage{siunitx} 
\usepackage[accsupp]{axessibility}

\usepackage[final]{iccv}      


\definecolor{iccvblue}{rgb}{0.21,0.49,0.74}
\usepackage[pagebackref,breaklinks,colorlinks,allcolors=iccvblue]{hyperref}

\def\paperID{9380} 
\def\confName{ICCV}
\def\confYear{2025}

\title{Self-Supervised Monocular 4D Scene Reconstruction for Egocentric Videos}

\author{\textbf{Chengbo Yuan\textsuperscript{1,2}, Geng Chen\textsuperscript{2,3}$^{\dag}$, Li Yi\textsuperscript{1,2}, Yang Gao\textsuperscript{1,2}$^\ddag$}\\
\textsuperscript{1}Institute for Interdisciplinary Information Sciences, Tsinghua University\\
\textsuperscript{2}Shanghai Qi Zhi Institute \ \ 
\textsuperscript{3}UC San Diego\\
{\tt\small ycb24@mails.tsinghua.edu.cn \ gec001@ucsd.edu} \\
{\tt\small \{ericyi, gaoyangiiis\}@mail.tsinghua.edu.cn}
}

\begin{document}
\maketitle

\renewcommand{\thefootnote}{}
\footnotetext{\dag\ Work done during the internship at Shanghai Qi Zhi Institute. \ddag\ The corresponding author.}

\begin{abstract}
Egocentric videos provide valuable insights into human interactions with the physical world, which has sparked growing interest in the computer vision and robotics communities. A critical challenge in fully understanding the geometry and dynamics of egocentric videos is dense scene reconstruction. However, the lack of high-quality labeled datasets in this field has hindered the effectiveness of current supervised learning methods. In this work, we aim to address this issue by exploring an self-supervised dynamic scene reconstruction approach. We introduce \textbf{EgoMono4D}, a novel model that unifies the estimation of multiple variables necessary for \underline{Ego}centric \underline{Mono}cular \underline{4D} reconstruction, including camera intrinsic, camera poses, and video depth, all within a fast feed-forward framework. Starting from pretrained single-frame depth and intrinsic estimation model, we extend it with camera poses estimation and align multi-frame results on large-scale unlabeled egocentric videos. 
We evaluate EgoMono4D in both in-domain and zero-shot generalization settings, achieving superior performance in dense pointclouds sequence reconstruction compared to all baselines. EgoMono4D represents the first attempt to apply self-supervised learning for pointclouds sequence reconstruction to the label-scarce egocentric field, enabling fast, dense, and generalizable reconstruction. The interactable visualization, code and trained models are released \href{https://egomono4d.github.io/}{https://egomono4d.github.io/}.

\end{abstract}

\section{Introduction}

Egocentric videos, especially Hand Object Interaction (HOI) videos, capture a vast amount of knowledge about human interaction with the physical world, particularly in tool usage. 
Due to this rich source of interaction knowledge, egocentric human videos have gained increasing interest from both the research community (e.g., computer vision \cite{egohos,egovlp} and robotics \cite{r3m,general_flow}) and industry (e.g., virtual reality \cite{adt}).

\begin{figure}[t]
  \centering
   \includegraphics[width=1.0\linewidth]{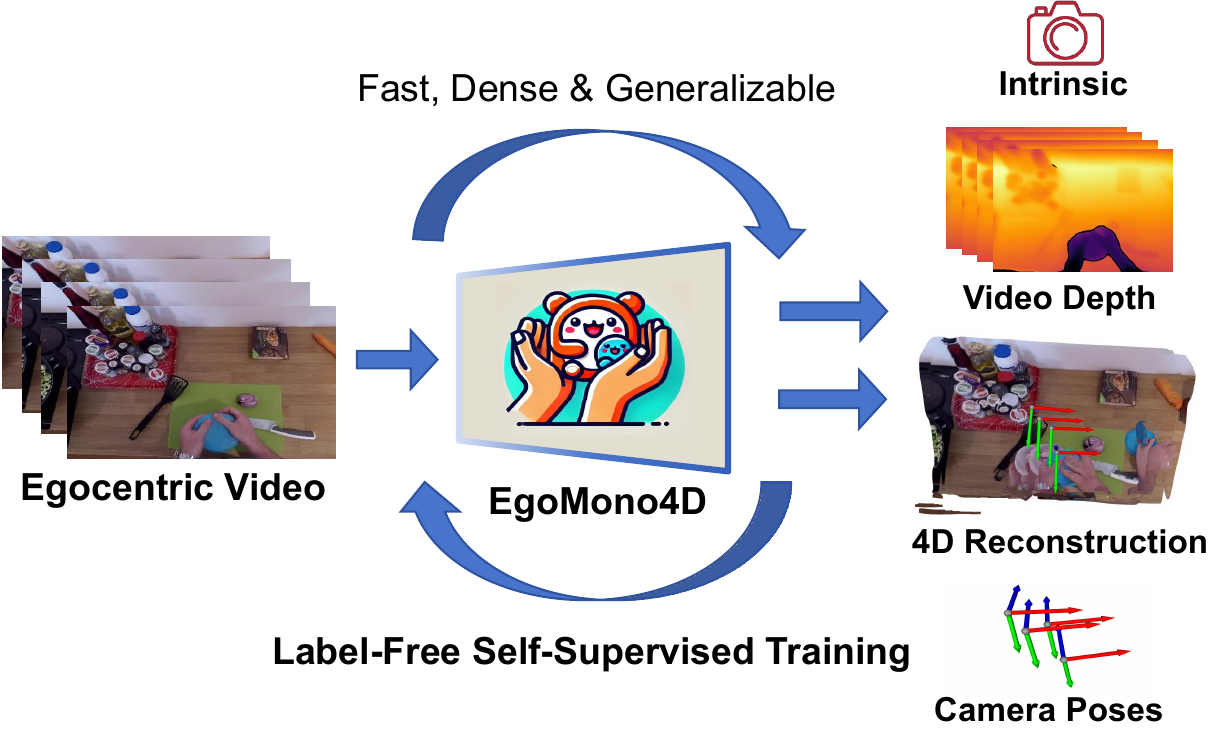}
   \caption{We propose EgoMono4D, a model that unifies the estimation of camera intrinsic, camera poses, and video depth for fast and dense 4D reconstruction of egocentric scenes. EgoMono4D is trained solely on large-scale unlabeled videos in an self-supervised learning framework.}
   \label{fig:1_intro}
\end{figure}

To better understand the geometry and dynamics in these activities, a crucial task is the dense 4D reconstruction from internet-scale egocentric video datasets \cite{epic_kitchen, ego4d}. 
Here, dense 4D reconstruction is represented as \textbf{a pointclouds sequence which captures the 3D position of every pixel in each frame within a global coordinate system} \cite{dust3r, monst3r, align3r, cut3r}, preserving maximal geometry details. 
This task requires: (1) dense per-pixel reconstruction, (2) the ability to handle dynamic motions in egocentric videos. To facilitate large-scale usage \cite{epic_kitchen, atm, general_flow}, (3) strong generalization and fast processing speeds are also needed to adapt to large-scale unseen egocentric scenes. 

However, current methods fail to meet these demands. Traditional methods \cite{sfm, droid_slam}, such as Structure-from-Motion (SfM) or SLAM system combined with dense depth estimation \cite{unidepth, metric3d}, face significant challenges in reconstructing dynamic scenes. These multi-step approaches also suffer from accumulated errors and inconsistencies between different modules \cite{dust3r,vggsfm}. Test-time optimization techniques~\cite{ego_gaussian, rcvd, mosca, dynamic_gaussion}, such as Gaussian Splatting \cite{gaussion_splatting}, suffer from slow processing speeds, making them impractical for reconstructing large-scale egocentric datasets \cite{ego4d, hoi4d}. Meanwhile, recent supervised learning approaches, such as DUSt3R~\cite{dust3r} and MonSt3R~\cite{monst3r}, are limited by the scarcity of labeled egocentric videos, especially outside of controlled environments \cite{epic_kitchen, ego4d}.

To address these challenges, we explore an self-supervised learning approach for fast, dense, and generalizable reconstruction of highly dynamic egocentric videos. We propose \textbf{EgoMono4D}, a model trained without ground-truth labels that simultaneously predicts multiple variables necessary for dense 4D reconstruction, including camera intrinsic, camera poses, and video depth \cite{flowmap, joint_prediction}.

Our key insight is to \textbf{extend pretrained single-frame scene reconstruction model to video version, and align multi-frame results with 4D constraints.} We begin by estimating per-frame depth and camera intrinsic to generate per-frame pointclouds predictions. Then, we align multi-frame results and derive camera extrinsics by minimizing the difference between (1) the 3D scene flow induced by the camera's motion through a static scene and (2) pre-computed 3D correspondences, similar to previous self-supervised methods \cite{unsupervised_depth, kick, kick++, flowmap}. A confidence mask is used to exclude dynamic and unreliable areas during multi-frame alignment. To prevent model collapse and accelerate training convergence, we also regularize the model with predictions from state-of-the-art off-the-shelf models, such as Unidepth~\cite{unidepth} for depth estimation and EgoHOS~\cite{egohos} for confidence mask prediction.

We evaluate EgoMono4D in both in-domain and zero-shot settings on unseen egocentric scenes. The model successfully recovers the 3D structure of scenes and the motion of dynamic parts, even in challenging synthetic surgery HOI videos \cite{pov_surgery}. We also provide a quantitative comparison with baseline methods that offer near-linear time complexity relative to the number of frames, focusing on two fundamental 4D tasks: dense pointclouds sequence reconstruction \cite{flowmap, dust3r} and long-term 3D scene flow recovery \cite{tapvid3d, general_flow, 3d_hand_traj}. EgoMono4D demonstrates superior performance on evaluation metrics, outperforming all baseline methods.

In conclusion, our main contributions are as follows:

\begin{itemize}
   \item We propose EgoMono4D, a model that unifies camera intrinsic, camera poses, and video depth estimation for fast, dense, and generalizable 4D reconstruction of egocentric videos.
   \item We introduce a novel self-supervised training method for egocentric scene reconstruction, training our model solely on large-scale, unlabeled monocular egocentric datasets, addressing the challenge of labeled data scarcity.
   \item EgoMono4D demonstrates promising performance in both in-domain and zero-shot unseen scenes, surpassing all baselines in pointclouds sequence reconstruction.
\end{itemize}

\begin{figure*}[t]
  \centering
   \includegraphics[width=0.95\linewidth]{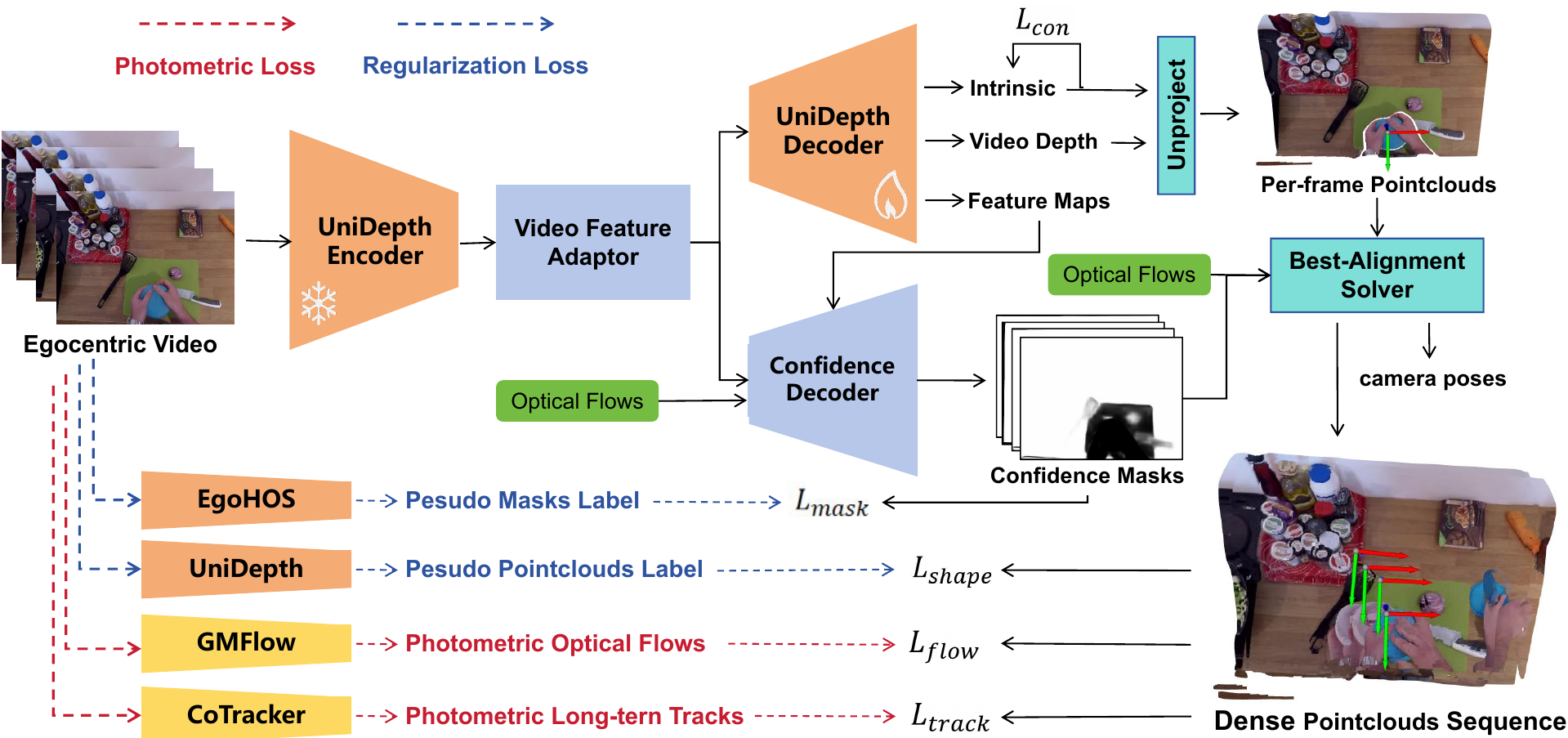}
   \caption{The overview of EgoMono4D and our self-supervised training framework. The model first simultaneously predicts camera intrinsic, video depth, and confidence maps  (for camera pose estimation). Camera poses are then calculated by aligning unprojected pointclouds from different frames with confidence maps. The final dense pointclouds sequence reconstruction is assembled using all the predicted variables. We train our model purely on unlabeled egocentric video datasets, with both self-supervised photometric loss for depth alignment and regularization loss for training stabilization. }
   \label{fig:2_framework}
\end{figure*}

\section{Related Works}

\subsection{(Self-Supervised) Monocular Depth Estimation}

Monocular depth estimation has made significant progress in recent years \cite{depth_survey}. Supervised learning models have shown strong generalization capabilities \cite{midas, zeodepth, depth_anything, metric3d, unidepth, depthcrafter, nvds}. Our work builds on UniDepth \cite{unidepth}, a state-of-the-art model that unifies camera intrinsic and depth estimation for single image.
Another line of works focus on self-supervised training \cite{joint_prediction, unsupervised_depth, sc_depth, swindepth, dynamo_depth}. These methods train depth estimators purely on monocular videos using photometric error supervision \cite{kick, kick++}, often by leveraging camera labels or learning camera predictors. Despite recent progress, no methods have yet demonstrated strong generalization across both camera and depth. Our approach share similar intuition, which also unifies depth and camera prediction and trained solely on monocular video datasets \cite{epic_kitchen} with photometric loss. 
However, our primary focus is on generalizable 4D reconstruction for egocentric scenes, which requires zero-shot prediction for both depth and camera parameters. To some extent, our work is the first attempt to extend self-supervised depth estimation to generalizable dense pointclouds sequence reconstruction.

\subsection{Structure from Motion and SLAM Systems}

Structure from Motion (SfM) \cite{sfm, sfm_global, sfm_pixel, particlesfm} and monocular visual SLAM systems \cite{orb_slam, orb_slam3, droid_slam, nerf-slam, go-slam} reconstruct 3D structures and estimate camera poses from image sequences. However, they struggle with dynamic scenes which pose ill condition for the epipolar constraint \cite{dust3r}. Moreover, most of them typically can not provide dense pixel-level reconstructions for all frames. Combining visual odometry \cite{droid_slam, tartanvo, leap-vo, mapfree_vr} with dense depth estimation \cite{unidepth, metric3d} helps constrain dynamic parts' geometry but can lead to accumulated and inconsistencies errors. Recently, FlowMap \cite{flowmap} offers a differentiable SfM for static scenes, optimizing depths, poses, and intrinsic simultaneously. We adopt its key ideas and extend it to a generalizable version for dynamic egocentric videos.

\subsection{Dense 4D Reconstruction for Dynamic Scenes}

Reconstructing 4D dynamic scenes remains a challenging problem in computer vision. Some approaches \cite{rcvd, causalsam, mosca, dynamic_gaussion, shape-of-motion, dreamscene4d, ego_gaussian, 3dgstream, dy_survey} first compute vision cues (e.g., camera pose, depth, optical flow) and then perform test-time optimization for each scene. While effective, these methods are time-consuming and impractical for large-scale reconstructions \cite{tapvid3d}.

Concurrent with our work, 
another line of research \cite{track-to-4d, dust3r, monst3r, align3r, megasam, mast3r_slam, mast3r-sfm, light3r, cut3r, stereo4d, btimer, spann3r} explores fast, feed-forward 4D reconstruction through supervised learning. Many of these methods are the extension of the pretrained end-to-end Multi-View Stereo (MVS) model DUSt3R \cite{dust3r}. For example, MonSt3R \cite{monst3r} and Stereo4D \cite{stereo4d} fine-tune DUSt3R for dynamic scenes, while Spann3R \cite{spann3r} and CUT3R \cite{cut3r} incorporate memory buffers for temporal information merging to avoid post-optimization. Align3R \cite{align3r} further enhances geometry estimation by merging depth priors. However, these methods rely on ground-truth labels for training, which are scarce for egocentric video datasets \cite{h2o}.
In this work, we aim to explore self-supervised methods \cite{flowmap, kick} for monocular egocentric scene reconstruction, leveraging large-scale unlabeled video datasets \cite{epic_kitchen} instead.

\subsection{Egocentric Video Understanding}

Egocentric videos and datasets \cite{fpha, egopat3d, epic_kitchen, adt, hoi4d, h2o, pov_surgery, arctic, ego4d} capture human hand interactions with the environment, which are critical for robotics \cite{r3m, general_flow, atm, vrb}, virtual reality \cite{adt}, and intelligent agents \cite{embodiedgpt, ego_llm}. Tasks like detection \cite{100doh}, segmentation \cite{egohos}, intention recognition \cite{egopat3d, affordancellm}, motion prediction \cite{joint_hand, 3d_hand_traj}, and hand-object reconstruction \cite{hert, wilor} are commonly used for egocentric videos. However, dense reconstruction of entire scenes remains a challenge. EgoGaussian \cite{ego_gaussian} solves this with time-intensive Gaussian Splatting \cite{gaussion_splatting}, while our method provides a faster ($>$ 30x speedup), feed-forward, and scalable solution.

\section{Preliminary}

\subsection{Problem Definition} \label{problem_definition}

A monocular egocentric video can be represented as a sequence of frames $\{I_t \in R^{H\times W\times 3}\}$, where $t=0,1...T-1$ and $T$ is the total number of frames. Each frame is a RGB image with a resolution of $H \times W$. Given a video, our goal is to estimate a sequence of dense pointclouds~\cite{dust3r}, $\{\hat{S}_{t}\in R^{H\times W\times 3}\}$, which capture the 3D position of every pixel in each frame within a global coordinate system. To achieve this, we decompose the pointclouds sequence into: 
(1) video depth $\{\hat{D}_{t} \in R^{H\times W}\}$; (2) camera intrinsic $\hat{K}\in R^{3\times 3}$ for unprojecting depth into per-frame pointclouds; and (3) camera poses $\{\hat{P}_{t}\in R^{4\times 4}\}$ in camera-to-world format for projecting the per-frame pointclouds into global coordinates. Then the pointclouds sequence $\{\hat{S}_{t}\}$ could be calculated as:
\begin{align}
& \hat{X}_t = \hat{D}_{t}(i,j) \hat{K}^{-1} h(p(i,j)) \label{eq:eq_1} \\
& \hat{S}_{t}(i,j) = h^{-1}(\hat{P}_{0}^{-1} \hat{P}_{t} h(\hat{X}_t))
\end{align}
where $(i,j)$ refers to the pixel position $p(i,j)$, $\hat{X}_t\in R^{H\times W\times 3}$ represents the per-frame pointclouds, and $h(\cdot)$ is the homogeneous operator that adds an extra dimension with a value of 1 to the coordinate system.

\subsection{Camera Pose from Depth Alignment} \label{sec: model_extrinsic}

Following FlowCam \cite{flowcam}, we \textbf{reframe camera pose estimation as a depth alignment and confidence mask prediction problem.} This approach transforms the task of predicting sparse camera parameters into \textbf{a dense pixel-level prediction problem}, enhancing robustness and generalization \cite{flowmap}. Specifically, the model first predicts multi-frame depth $\hat{D}_t$ and camera intrinsic $\hat{K}$, and then unprojects the depths into per-frame pointcloudss $\hat{X}_t$. We use an off-the-shelf model \cite{gmflow} to compute the optical flow between adjacent frames, denoted as $\hat{u}_{i-1,i}$. The optimal camera pose transformation should best align the 3D point pairs induced by $\hat{u}_{i-1,i}$. Additionally, we predict a confidence mask $\hat{\mathcal{M}}_{i,i-1}$ for frame $i$ to exclude (1) dynamic regions, (2) occlusions, (3) scene edges, and (4) inaccuracies in optical flow during the alignment process.

Formally, let $\hat{X}_{i}^{\leftarrow i-1}$ denote the result of interpolating $\hat{X}_{i}$ using the points from $\hat{X}_{i-1}$ (this can be computed based on $\hat{u}_{i-1,i}$). The best-aligned camera pose transformation $\hat{P}_{i,i-1}$ can then be formulated as:

\begin{equation} \label{eq: procrutes}
\hat{P}_{i,i-1} =  \mathop{\arg\min}\limits_{P\in SE(3)}  ||\hat{\mathcal{M}}_{i,i-1} (\hat{X}_{i-1} - P \hat{X}_{i}^{\leftarrow i-1})|| 
\end{equation} 
Then transformation $\hat{P}_{i,j}$ between arbitray frames $i$ and $j$ could be acquire by chain the nearby-frame results. Solving Equation~\ref{eq: procrutes} is known as the weighted procrustes-alignment problem~\cite{flowcam}, which can be solved in closed form using the singular value decomposition (SVD) \cite{svd}, allowing a differentiable and learnable camera pose estimation process via gradient descent \cite{flowmap}. This means that we could get camera pose naturally by only focus on predicting video depth, camera intrinsic and confidence masks.

\section{Methodology} 

\subsection{Overview}

We begin with the state-of-the-art pretrained single-frame scene reconstruction model, UniDepth \cite{unidepth}, which predicts single-frame depth and intrinsic. Our goal is to extend it to a video-based model with label-free training. To achieve this, we need to (1) predict camera poses and (2) eliminate inconsistencies between multi-frame results.

To enable camera poses estimation, we adapt UniDepth from an image to a video estimator using adaptor blocks \cite{fs_adaptor, vit_adaptor}. We also introduce a new decoder to predict confidence masks, facilitating camera poses estimation as described in Section~\ref{sec: model_extrinsic}. To eliminate multi-frame inconsistencies, we employ self-supervised training losses based on 4D constraints to align multi-frame results, ensuring both temporal and spatial consistency in video predictions. The detailed approach is outlined in the following section.

\subsection{Model Architecture} \label{model_architecture}

Our model aims to predict (1) video depth, (2) camera intrinsic, and (3) confidence masks. Camera poses are then derivated using multi-frame depth alignment following Section~\ref{sec: model_extrinsic}. Our model builds upon UniDepth \cite{unidepth}, a universal estimator for predicting single-frame depth and camera intrinsic with an \textbf{encoder-decoder} architecture. More details about UniDepth can be found in Appendix~\ref{app: unidepth}. We adopt its encoder and decoder for image encoding and depth and intrinsic prediction.
To adapt to 4D video reconstruction, we introduce two modifications to the UniDepth backbone:

\textbf{From Image to Video Estimation}: To facilitate video prediction, we use adaptor blocks\cite{vit_adaptor, fs_adaptor} to extend the original image estimator to video version.
Our model processes $N_w$ input images, extracting features from each frame individually through the encoder. The UniDepth encoder produces two types of features: (1) DINO \cite{dinov2} features $F_{dino}\in R^{T\times \frac{H}{s_h} \times \frac{W}{s_w} \times D_{dino}}$, where $s_h \times s_w$ represents the patch size in DINO and $D_{dino}$ is the feature dimension, and (2) global token features $F_{global}\in R^{T\times D_{global}}$. To fuse the features across time, we incorporate multiple adaptors \cite{vit_adaptor}. Global token features are fused using a Transformer \cite{attention} on temporal dimension, while the patched DINO \cite{dinov2} features are fused using Unet3D \cite{unet3d} on both temporal and spatial dimension. The architecture for the depth and intrinsic decoders remains unchanged from the original UniDepth implementation.

\textbf{New Confidence Mask Decoder}: To enable camera poses derivation, we need to predict an extra confidence mask as mentioned in Section~\ref{sec: model_extrinsic}. We add a new confidence mask decoder adopted from \cite{flowmap}, which is a 3-layer MLP with ReLU \cite{relu} activation. The decoder takes a concatenation of (1) fused shallow features from the video adaptors, (2) depth features and confidence maps from the UniDepth decoder \cite{unidepth}, and (3) an interpolation of the above features, induced by optical flow \cite{gmflow} from neighboring frames (Secntion~\ref{sec: model_extrinsic}). A sigmoid function is applied to normalize the final confidence score within the range $[0,1]$.

\subsection{Self-supervised 4D Reconstruction Losses} \label{sec: model_training}

Although the new architecture can predict camera intrinsics, poses, and video depth simultaneously, these variants remain inconsistent in both the temporal and spatial dimensions. To address these issues, we propose an self-supervised training method that optimizes and aligns these variants in an end-to-end manner. By leveraging several 4D geometric constraints, we design self-supervised training losses to enable label-free training. We categorize our losses into two types: (1) Photometric loss, which aligns depth, intrinsics, and extrinsics to ensure consistent 4D reconstruction, and (2) Regularization loss, which accelerates training convergence and helps prevent model collapse.

\subsubsection{Photometric Loss from Flow and Track Prior}

Similar to previous methods \cite{unsupervised_depth, kick, kick++, flowcam, monst3r, align3r}, we use photometric loss to align multi-frame depth estimations. This also ensures consistency between the camera parameters, poses and depth predictions.
Specifically, we first back-project depth and intrinsic into per-frame pointcloudss $\hat{X}_i$. Then, we align the multi-frame results by minimizing the difference between (1) the 3D scene flow and long-term tracking induced by the camera’s movement through high-confidence areas and (2) pre-computed 3D correspondences (back-projected from the optical flow computed by GMFlow \cite{gmflow} and long-term tracking by CoTracker \cite{cotracker}).

Formally, suppose $i < j$, and $\hat{X}_j^{\leftarrow i}$ is the interpolation result of $\hat{X}_j$ based on the points of $\hat{X}_i$ (which can be computed using optical flow or tracking). The alignment minimizes the 3D reprojection error in high-confidence regions between frames $i$ and $j$ in a \textbf{scale-agnostic} manner, which can be expressed as: 

\begin{equation}
\mathcal{L}_{flow/track} = \frac{||\tilde{M}_{i+1,i} \tilde{M}_{j,j-1} (\hat{X}_i - \hat{P}_{j,i} \hat{X}_j^{\leftarrow i})||}{F(\hat{X}_j)||\tilde{M}_{i+1,i}\tilde{M}_{j,j-1}||} 
\end{equation}
where $F(\cdot)$ computes the first principal component of the pointcloudss. We use $F(\hat{X}_i)$ as a proxy for the scale of $\hat{X}_{t}$ and place it in the denominator of $\mathcal{L}_{flow/track}$ to prevent the pointcloudss from collapsing to a single point. Compared to widely used 2D photometric loss \cite{kick, flowmap}, $\mathcal{L}_{flow/track}$ encourages more intuitive 3D consistency in the predictions.

Note we use a pre-computed pseudo-confidence mask $\tilde{M}$ for the photometric loss instead of the predicted mask $\hat{M}$, since the pseudo-motion mask is able to approximate from pretrained segmentation model~\cite{egohos} in egocentric video. The predicted mask $\hat{M}$ is optimized by backpropagating $\mathcal{L}_{flow/track}$ through $\hat{P}_{j,i}$, as described in Section~\ref{sec: model_extrinsic}. Using $\tilde{M}$ improves model stability and robustness by promoting more correspondences and preventing $\hat{M}$ from shrinking into sparse predictions. The pseudo-mask $\tilde{M}$ is computed from two sources: (1) Pseudo-dynamic areas, using a hand and interacted objects mask from EgoHOS \cite{egohos}, which captures motion from hand-object interactions; (2) Pseudo-edges, derived from flying pixels \cite{fly_pixel} based on UniDepth \cite{unidepth} depth predictions. For the Epic-Kitchen \cite{epic_kitchen} dataset, dynamic masks are also estimated using epipolar loss \cite{epipolar_loss} when hands are outside the camera view.

\subsubsection{Regularization Loss from Depth and HOI Prior}

To stabilize the model training process and accelerate convergence, we also regularize the training with predictions from state-of-the-art off-the-shelf models, i.e., Unidepth~\cite{unidepth} for depth estimation and EgoHOS~\cite{egohos} for confidence mask prediction.

\vspace{0.15cm}

\noindent \textbf{Shape Regularization Loss from Depth Prior}\ \ After predicting the depth ${\hat{D}_t}$ and camera intrinsic $\hat{K}$, we first recover per-frame pointclouds $\hat{X}_t$ using Equation~\eqref{eq:eq_1}. We then regularize the shape of $\hat{X}_{t}$ with the prediction $\widetilde{X}_t$ from UniDepth \cite{unidepth}: 
\begin{equation}
    \mathcal{L}_{shape} = \frac{1}{H\times W} \mathop{\min}\limits_{s, R, T}  ||sR\hat{X}_{t} + T - \widetilde{X}_t|| 
\end{equation}
Here, $(s, R, T)$ represents the scaled SE(3) transformation used for alignment, aiming to enable regularization on a relative scale \cite{midas}. The optimal transformation can be solved in closed form using SVD \cite{svd}. Note that $\mathcal{L}_{shape}$ helps constrain dynamic parts of scenes relative to static areas.

\vspace{0.15cm}

\noindent \textbf{Mask Regularization from HOI Prior}\ \ To speed up convergence, we also regularize the prediction of $\hat{M}$: 

\begin{equation}
\mathcal{L}_{mask} = BCELoss(\hat{\mathcal{M}}, \widetilde{\mathcal{M}})
\end{equation}

\vspace{0.15cm}

\noindent \textbf{Camera Consistency Self-Supervision}\ \ We additionally introduce a self-supervised loss $\mathcal{L}_{con}$ to enhance the consistency of camera intrinsic predictions. For two frame sequences $V_1$ and $V_2$ from the same video clip, we enforce the intrinsic predictions to be as similar as possible: \begin{equation} 
\mathcal{L}_{con} = || \hat{K}_{V_1} - \hat{K}_{V_2} || 
\end{equation}

\subsubsection{Final Loss Function}

Put it together, the final loss  is:
\begin{equation}
\mathcal{L} = \alpha \mathcal{L}_{shape} + \beta \mathcal{L}_{flow} + \gamma \mathcal{L}_{track} + \lambda \mathcal{L}_{mask} + \mu \mathcal{L}_{con}
\end{equation}

where we set $\alpha=4$, $\beta=\gamma=5$, $\lambda=1$, $\mu=0.005$ as loss weights by default to balance each supervision.  

\begin{figure}[t]
  \centering
   \includegraphics[width=1.0\linewidth]{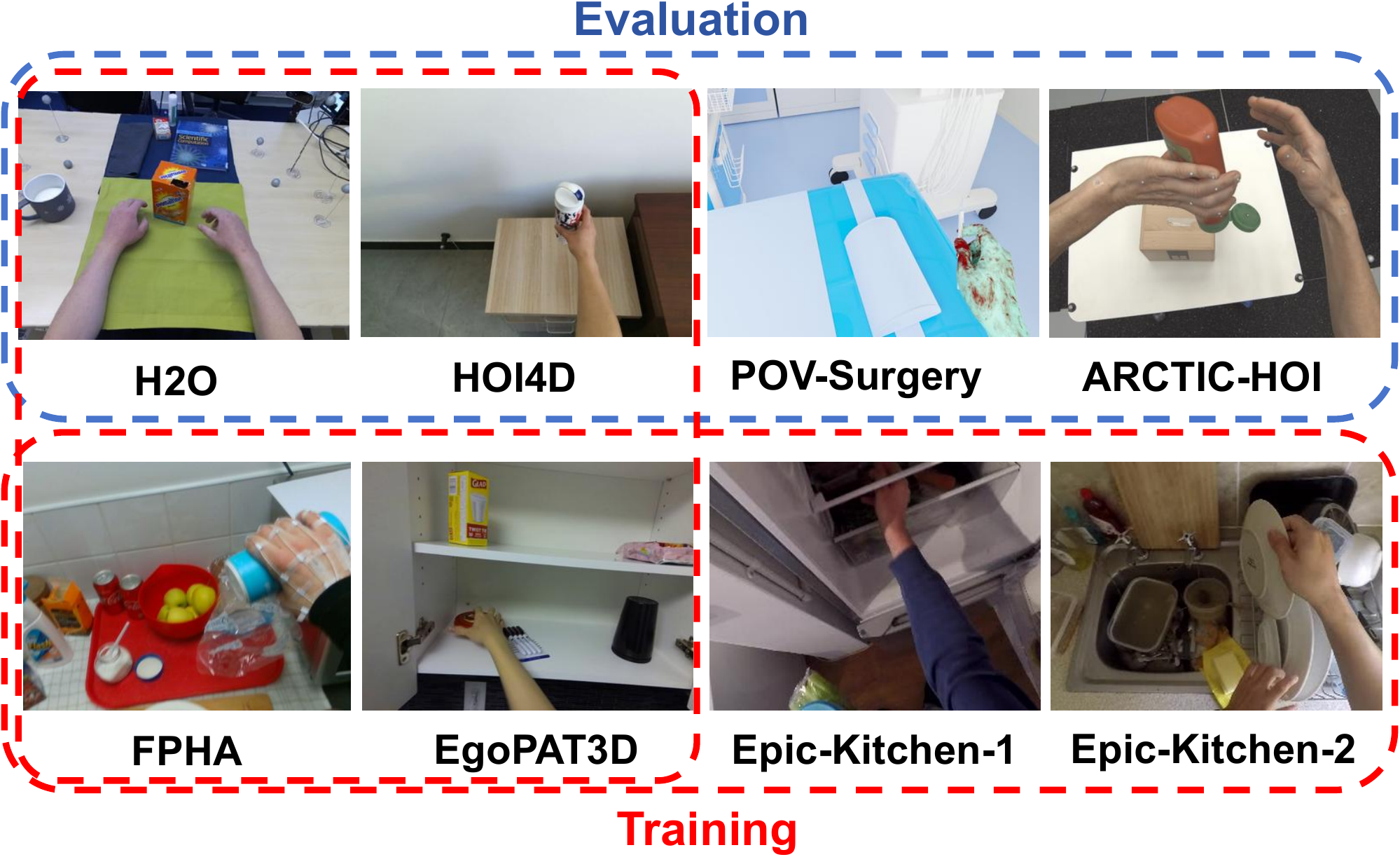}
   \caption{Visualization of dataset used for training and evaluation.}
   \label{fig:dataset}
\end{figure}

\subsection{Inference Strategy} \label{model_inference}

 Finally, we describe the inference strategy for EgoMono4D. Due to GPU memory limitations, only a limited number of frames can be processed in a single feed-forward prediction.
However, we can predict videos with infinite frames in a stream manner using a sliding window. The video is first split into $N_w$ frames sub-clips with $N_o$ overlapping frames between neighbors. Then we predict neighboring windows independently. Let $w_i$ represent the $i$-th sub-clip, $E_i$ the timestamp set of $w_i$, and $\hat{S}^{w_i}$, $\hat{S}^{w_{i+1}}$ the predictions for two neighboring sub-clips. The latter is then then aligned and concatenated to the former as follows:
\begin{align}
&(s^{*},R^{*},T^{*}) =  \mathop{\arg\min}\limits_{s, R, T} ||sR\hat{S}^{w_i}_{E^{ov}_{i+1}} + T - \hat{S}^{w_{i+1}}_{E^{ov}_{i+1}}|| \\
& \hat{S}^{[w_i,w_{i+1}]} = [\hat{S}^{w_{i}}, s^{*}R^{*} \hat{S}^{w_{i+1}}_{E_{i+1} - E^{ov}_{i+1}} + T^{*}]
\end{align}
where $E^{ov}_{i+1}$ represents the overlapping timestamps between $w_i$ and $w_{i+1}$, and $[\cdot, \cdot]$ denotes the concatenation operator.

\begin{table*}[!t]
\centering
\small
\setlength{\tabcolsep}{3pt} 
\newcolumntype{C}{>{\centering\arraybackslash}X} 
\begin{tabularx}{\textwidth}{lCCCCCCCCCCCCCCCC}
\toprule
                      & \multicolumn{4}{c}{\textbf{HOI4D}}    & \multicolumn{4}{c}{\textbf{H2O}}   & \multicolumn{4}{c}{\textbf{POV-Surgery$^{\dag
}$}}  & \multicolumn{4}{c}{\textbf{ARCTIC-HOI$^{\dag
}$}} \\ 
                      \cmidrule(lr){2-5} \cmidrule(lr){6-9} \cmidrule(lr){10-13} \cmidrule(lr){14-17}
                      & {CD $\downarrow$} & {$F_{1} \uparrow$} & {$F_{2.5} \uparrow$} & {$F_{5} \uparrow$} & {CD $\downarrow$} & {$F_{1} \uparrow$} & {$F_{2.5} \uparrow$} & {$F_{5} \uparrow$} & {CD $\downarrow$} & {$F_{1} \uparrow$} &  {$F_{2.5} \uparrow$} & {$F_{5} \uparrow$} & {CD $\downarrow$} & {$F_{1} \uparrow$} & {$F_{2.5} \uparrow$} & {$F_{5} \uparrow$} \\
\midrule
Modularized Version  & 8.9  & 15.4 & 38.2  & 68.0  &\textbf{ 4.7} & \underline{47.8} & \underline{81.5}  & \underline{94.2} & 223.3 & 3.6 & 9.6 & 19.1 & 5.8  & 12.6 & 33.8 & 63.2    \\
DS+UniDepth~\cite{droid_slam} & \underline{6.7} & \underline{23.2}  & \underline{53.4}  & \underline{79.9}  & \underline{5.1} & 36.5 & 75.0  & 93.2 & \underline{39.1}  & 7.7 & 20.0  & 38.7 & \underline{2.9}  & \underline{22.2}  &\textbf{60.2}  & \underline{84.7}     \\
DUSt3R~\cite{dust3r}    & 8.6  & 24.0 & 53.2  & 76.8  & 8.8 & 23.1 & 56.0  & 82.7 & 55.4 &  \underline{9.7} & \underline{24.7} & \underline{45.1} & 3.2  & 20.8 & 53.7 & 83.1     \\
MonSt3R~\cite{monst3r}      &  7.6 & 21.4 & 50.8 & 77.9 & 14.7 & 15.1 & 42.4 & 70.0 & 94.5& 6.5 & 17.8 & 34.8  & 5.4 & 9.9 & 31.8 & 65.0    \\
Align3R~\cite{align3r}      &  7.1 & 22.3 & \underline{53.4} & 78.9 & 11.5 & 20.0 & 45.4 & 72.0 & 41.1 & 7.9 & 21.0 & 40.4  & 4.5 & 14.2 & 41.5 & 75.8    \\
CUT3R~\cite{cut3r}      & 7.5 & 20.1 & 47.5 & 74.3 & 7.4 & 17.2 & 57.2 & 93.1 & 107.6 & 4.7 & 12.8 & 25.6  & 4.5 & 18.7 & 48.2 & 77.9    \\
\midrule
EgoMono4D (Ours)     & \textbf{5.9}  & \textbf{27.9} & \textbf{59.6}  &  \textbf{83.1}   & \underline{5.1}  & \textbf{54.2} & \textbf{83.9}  &  \textbf{94.4} & \textbf{33.8} & \textbf{13.5} & \textbf{32.0} & \textbf{53.9} & \textbf{2.8}   & \textbf{24.1} & \underline{57.5}  & \textbf{86.2}     \\
\bottomrule
\end{tabularx}
\caption{The evaluation results for 4D pointclouds sequence reconstruction are presented, using 3D Chamfer Distance (CD, mm) and 3D Pointclouds F-score (F$_{\delta}$, \%). $\dag$ indicates zero-shot generalization for EgoMono4D. For ARCTIC-HOI, the evaluation focuses specifically on the reconstruction quality of the hand-object region. On average, EgoMono4D demonstrates a clear advantage across the metrics.}
\label{tab: pcd_reconstruction}
\end{table*}

\section{Experiments}

\subsection{Datasets}
Figure~\ref{fig:dataset} shows the datasets used for training and evaluation. For more details, refer to Appendix~\ref{app: datasets}. Our model is trained on egocentric videos from H2O\cite{h2o}, HOI4D\cite{hoi4d}, FPHA\cite{fpha}, EgoPAT3D\cite{egopat3d}, and Epic-Kitchen\cite{epic_kitchen}. Each video is split into 20-frame sub-clips for batch training, totaling 11.2 million frames, with the majority (9.7M frames) from the unlabeled Epic-Kitchen dataset.

For evaluation, we use datasets with pointcloud sequence labels. In-domain evaluation is done using H2O\cite{h2o} and HOI4D\cite{hoi4d}, with the datasets split by Scene ID to ensure no overlap between training and test sets. For zero-shot generalization, we use POV-Surgery\cite{pov_surgery} and ARCTIC\cite{arctic} (only Mocap \cite{mocap} Hand and Objects Interaction (HOI) labels). To avoid redundancy, we only use the first record from the first participant in each task. Videos are split into 40-frame sub-clips for batch evaluation. Note that ARCTIC provides only hand and object labels, so we refer to it as ARCTIC-HOI to highlight its focus on foreground HOI reconstruction.

\begin{figure*}[t]
  \centering
   \includegraphics[width=1.0\linewidth]{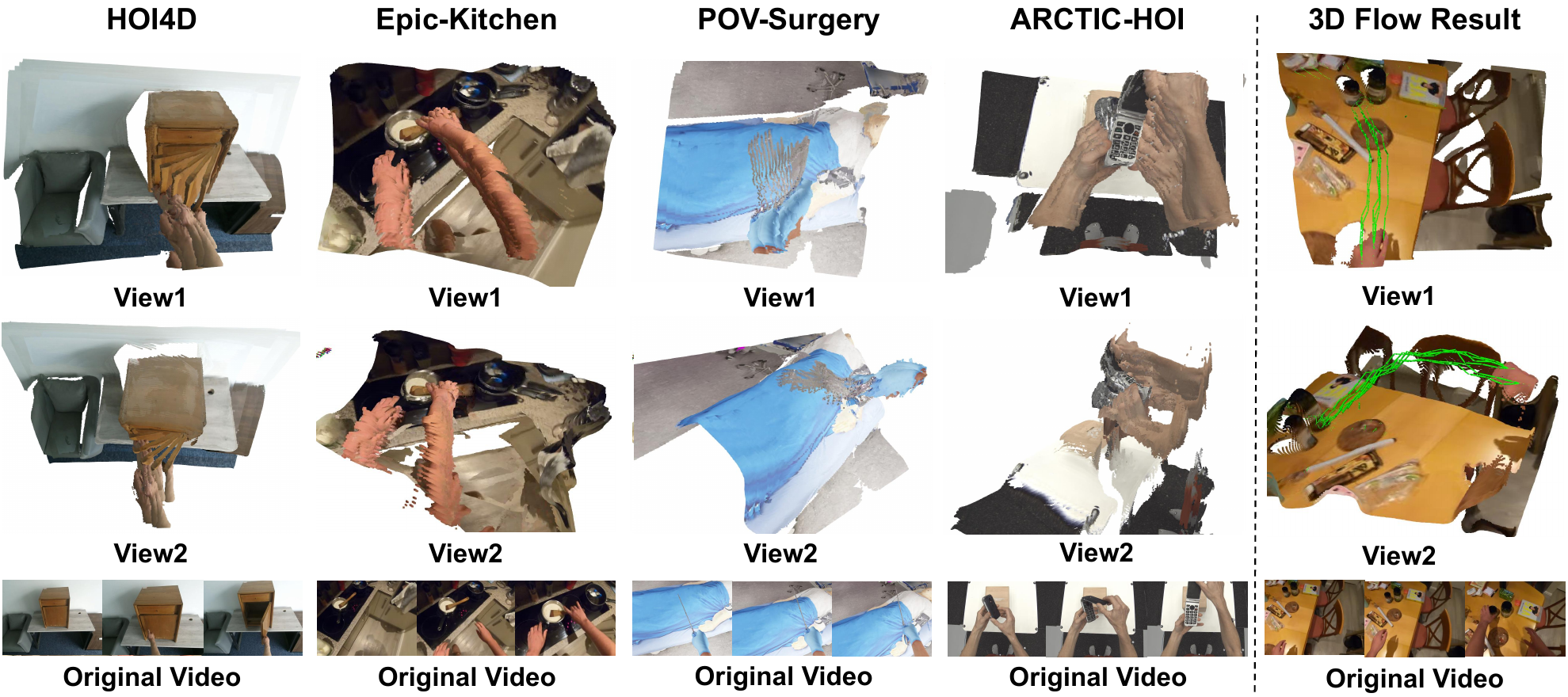}
   \caption{
   The visualization of the dense pointcloud sequence reconstruction by EgoMono4D demonstrates its ability to effectively recover both the overall scene structure and dynamic motion elements to a significant extent. For additional visualizations, please refer to Appendix~\ref{app: more_vis}. We also provide a \textbf{qualitative comparison} and visualization with baseline methods in Appendix~\ref{app: qualitative_comp}. Video visualizations and interactive demonstration could be found in \href{https://egomono4d.github.io/}{https://egomono4d.github.io/}.
   }
   \label{fig:3_pcd_visualization}
\end{figure*}

\subsection{Implementation Details}

We initialize our model with the pretrained UniDepthV2-L weights\cite{unidepth} and freeze the encoder. For input preprocessing, we resize the images to a resolution of 288 $\times$ 384. During training, each data point consists of 4 frames sampled from each sub-clip, with the interval between frames randomly selected from the range $[1,4]$. We employ the Adam \cite{adam} optimizer with a learning rate of 5e-5 and a batch size of 16. The model is trained on 8 NVIDIA A800 GPUs for 350k iterations. For inference, we set the overlap to 1 frame. For stability, we set window size $N_w=4$ by default, the same with training process. 

\subsection{Baseline}

We compare our model to previous methods that (1) provide dense 4D reconstruction, (2) have nearly linear time complexity with respect to the number of frames, enabling large-scale reconstruction, and (3) demonstrate zero-shot generalization. \textbf{Modularized Version} (MapFreeVR \cite{mapfree_vr} + UniDepth \cite{unidepth}) first estimates image depth, then computes camera poses with correspondence estimation (optical flow \cite{gmflow} in our setting) and depth-alignment. Additionally, we integrate HOI masks from EgoHOS\cite{egohos} to filter out the HOI region for alignment. This baseline could be viewed as a modularized and no-training version of EgoMono4D. \textbf{DS+UniDepth} (DROID-SLAM \cite{droid_slam} + UniDepth \cite{unidepth}) combines depth estimation with learning-based RGBD visual odometry. \textbf{DUSt3R} \cite{dust3r} supports end-to-end reconstruction. We use the "swin" mode from the original implementation to achieve $\mathcal{O}(T)$ inference. \textbf{MonSt3R} \cite{monst3r} is a finetuned version of DUSt3R on synthetic dynamic datasets, aimed at improving reconstruction performance in dynamic areas. \textbf{Align3R} \cite{align3r} further merges depth prior during finetuning to enhance the geometry estimation ability. Lastly, \textbf{CUT3R} \cite{cut3r} incorporates memory buffers for temporal information merging to improve reconstruction consistency.

Since large-scale labeled 4D datasets (with both high-quality depth and camera labels) for in-the-wild egocentric videos are lacking, existing evaluation are limited to small, lab-based settings \cite{hoi4d, h2o, pov_surgery, arctic}. Fine-tuning on these specialized datasets could lead to overfitting, resulting in an unfair comparison, since EgoMono4D is trained on large-scale, in-the-wild datasets and designed for generalization. To avoid this, and following previous self-supervised depth estimation works \cite{unsupervised_depth, kick, kick++}, we refrain from fine-tuning models on lab-based egocentric datasets. Moreover, our aim is to evaluate how self-supervised methods perform in label-scarce settings, such as egocentric videos. By excluding labels, we gain clearer insights into how these methods tackle the inherent challenges of such domains. 

\section{Result}

Here we present the evaluation and ablation results of EgoMono4D. Detailed definitions of all metrics can be found in Appendix~\ref{app: metric}. We include the speed measurement and comparison of all models in Appendix~\ref{app: inference_speed}. We also include the evaluation results of depth and camera poses estimation in Appendix~\ref{app: depth_camera}.

\subsection{Dense Pointclouds Sequence Reconstruction}

\noindent \textbf{Task and Metric}\ \ Dense pointclouds sequence reconstruction \cite{flowmap, dust3r, monst3r} requires the model to reconstruct the $H \times W$ pointclouds of each frame within a global coordinate system. Since the scale of the pointclouds is ambiguous for monocular reconstruction, we first apply an estimated globally scaled SE(3) transformation $(s, R, T)$ to align the prediction to the ground-truth. We then evaluate the accuracy of each frame’s shape and average these results as the final evaluation score. Following \cite{pcd_metric, unidepth}, we use the 3D Chamfer Distance (CD, mm) and the 3D Pointclouds F-score (F$_{\delta}$, \%) to assess shape similarity. We use the notation F$_{\delta}$ to denote the F-score with a threshold of $\delta$ centimeters (cm) as the positive criterion.

\vspace{0.25cm}

\noindent \textbf{Results}\ \ The quantitative evaluation results are presented in Table~\ref{tab: pcd_reconstruction}, with visualizations shown in Figure~\ref{fig:3_pcd_visualization}. For additional visualizations, refer to Appendix~\ref{app: more_vis}. We also provide a \textbf{qualitative comparison} and visualization with baseline methods in Appendix~\ref{app: qualitative_comp}. Video and interactive demonstration visualizations can be found in \href{https://egomono4d.github.io/}{project website}.

EgoMono4D demonstrates superior performance across all evaluated methods. Notably, on the challenging POV-Surgery dataset \cite{pov_surgery}, which contains complex surgical scenes with unrealistic textures and intricate actions, our model outperforms others by 10–20\% in terms of F-score. DS+UniDepth also shows commendable performance across several metrics. However, qualitative visualizations (Appendix~\ref{app: qualitative_comp}) reveal that it struggles to align the static portions of egocentric scenes.

The modularized version of our model performs significantly worse across most metrics. This can be attributed to inconsistencies between the different modules. In contrast, our end-to-end self-supervised training (described in Section~\ref{sec: model_training}) ensures that the modules align through back-propagation of the 4D supervision loss, leading to improved 4D reconstruction accuracy.

Due to the limited availability of labeled egocentric data for training, DUSt3R, MonSt3R, Align3R and CUT3R exhibit suboptimal performance on egocentric videos. Furthermore, fine-tuning MonSt3R on small-scale dynamic datasets actually results in performance degradation compared to DUSt3R, primarily due to domain gap and overfitting. By introducing depth prior, Align3R alleviates this issue, but still fails to get precise monocular 4D estimation.
These results highlight the challenges that supervised learning methods face in label-scarce scenarios.  In contrast, our self-supervised approach performs effectively on unlabeled egocentric videos. It is important to note that this does not imply that supervised methods are inherently inferior to self-supervised ones. In domains with abundant labeled data, supervised methods may offer advantages, as demonstrated in depth estimation tasks \cite{depth_survey}.

\subsection{Long-term 3D Scene Flow Recovery}

We evaluate different models using long-term 3D scene flow \cite{3d_scene_flow}, which captures both the structure and dynamics of egocentric scenes. 
Given a video and query points in the first frame, long-term 3D flow represents the future trajectory of each point in 3D space \cite{general_flow, tapvid3d}. EgoMono4D also outperforms all baseline in this task. Details and results are in Appendix~\ref{app: flow_result}, with an prediction example in Figure~\ref{fig:3_pcd_visualization}.

\begin{table}[!t]
\centering
\small
\renewcommand{\arraystretch}{0.85}
\begin{tabular}{lcccc}
\toprule
          & \textbf{CD} & \textbf{F$_{1}$} & \textbf{F$_{2.5}$} & \textbf{F$_{5}$} \\ 
\midrule
complete                      & 6.3     & 26.9       & 56.9            & 81.4                  \\
\midrule
w UniDepth-S                     & 6.5 $\downarrow$     & 23.8 $\downarrow$    & 54.5 $\downarrow$                       & 79.2 $\downarrow$          \\
w/o V-Adaptor        & 6.4 $\downarrow$     & 25.8 $\downarrow$    & 55.8 $\downarrow$                       & 80.4 $\downarrow$          \\
w $\alpha=1$        & 7.1 $\downarrow$     & 18.2 $\downarrow$    & 48.3 $\downarrow$                       & 66.4 $\downarrow$          \\
w/ midas-loss                     & 6.3 $\cdot$     & 26.1 $\downarrow$    & 56.2 $\downarrow$                       & 80.4 $\downarrow$          \\
w/ 2d-flow-loss                    & 6.2 $\uparrow$     & 26.7 $\downarrow$        & 57.0 $\uparrow$                   & 81.0 $\downarrow$          \\
w/o mask-loss               & 6.5 $\downarrow$       & 23.9 $\downarrow$     & 53.3 $\downarrow$            & 79.0 $\downarrow$                  \\
w/o cc-loss                        & 6.3 $\cdot$         & 25.2 $\downarrow$     & 56.4 $\downarrow$            & 81.1 $\downarrow$                \\

\bottomrule
\end{tabular}
\caption{Ablation study of the EgoMono4D model on the HOI4D dataset. $\downarrow$ indicates performance degradation relative to the complete model, while $\uparrow$ indicates improvement. The complete model outperforms other ablated variants on average.}
\label{tab: abl_ex}
\end{table}

\begin{table}[!t]
\centering
\small
\renewcommand{\arraystretch}{0.85}
\begin{tabular}{c|>{\centering\arraybackslash}p{1.2cm}>{\centering\arraybackslash}p{1.2cm}>{\centering\arraybackslash}p{1.2cm}>{\centering\arraybackslash}p{1.2cm}}
\toprule
\multicolumn{1}{l}{\multirow{2}{*}{}} & \multicolumn{4}{c}{\textbf{POV-Surgery}} \\ 
\cmidrule(lr){2-5} 
                                       & \textbf{CD$\downarrow$} & \textbf{F$_{1}\uparrow$} & \textbf{F$_{2.5}\uparrow$} & \textbf{F$_{5}\uparrow$}   \\ 
\midrule
fps / 1                                & 25.5     & 14.2    & 33.3            & 54.8              \\ 
fps / 2                                & 25.4            & 13.8    & 32.8            & 54.9               \\ 
fps / 4                                & 25.2       & 14.1    & 33.1            & 55.1                    \\ 
fps / 12                               & 28.5      & 13.0     & 31.0            & 53.5                    \\ 
\bottomrule
\end{tabular}
\caption{The POV-Surgery pointclouds sequence reconstruction results across different frames per second (fps) settings.}
\label{tab: fps_result}
\end{table}

\subsection{Ablation Study}

We conduct an ablation study of EgoMono4D, training the model on 280K frames from the training set and testing its performance on the HOI4D dataset~\cite{hoi4d}. The variants tested are as follows: (1) w UniDepth-S: using small version of UniDepth~\cite{unidepth}. (2) w/o V-Adaptor: remove video adaptor. (3) w $\alpha=1$, changing the training weight. (4) w/ midas-loss: replacing $\mathcal{L}_{shape}$ with depth supervision at a relative scale \cite{midas}; (5) 2d-flow-loss: substituting $\mathcal{L}_{flow/track}$ with its 2D version in pixel space \cite{flowmap}; (6) w/o mask-loss: removing the confidence mask regularization loss $\mathcal{L}_{mask}$; and (4) w/o cc-loss: removing the self-supervised intrinsic loss $\mathcal{L}_{cc}$.

Results are presented in Table~\ref{tab: abl_ex}. The complete model outperforms all other variants on average, demonstrating the effectiveness of our design choices. From the results, we observe that mask and depth regularization plays a crucial role in stabilizing training. Additionally, the choice of the base model is important, and we anticipate that improvements in depth estimation will further benefit our self-supervised scene reconstruction approach in the future. In terms of shape regularization and photometric loss, applying constraints in 3D space yields moderately better results than applying them in 2D space on average.

We also conduct an ablation on the hyperparameters for model inference, including the window size $N_w$ and window overlap size $N_o$, which are detailed in Appendix~\ref{app: infer_ablation}. For the window size $N_w$, maintaining consistency between training and inference is critical, likely because the video adaptor is specifically trained to fuse 4 frames. Regarding the overlap size $N_o$, the model performs comparably for $N_o = 1, 2, 3$. Therefore, we select $N_o = 1$ to maximize inference speed.

\subsection{Impact on Video FPS}

Since we use random frame intervals to sample data during training, our model is expected to be robust to variations in video fps to some extent. We test this on the pointclouds sequence reconstruction task using the zero-shot dataset POV-Surgery \cite{pov_surgery}. For a 40-frame sub-clip of POV-Surgery, we select only frames 0, 12, 24, and 36 for evaluation. We define a $1/x$ fps video as a frame sequence sampled with an interval of $x$ (where 12 should be divisible by $x$). The evaluation results are shown in Table~\ref{tab: fps_result}, and they demonstrate that our method is robust to changes in video fps within a certain range. However, when the fps becomes too low, performance degrades, likely because the optical flow module \cite{gmflow} we rely on may fail under such conditions. 

\section{Conclusion}

We present EgoMono4D, an self-supervised model for 4D reconstruction of egocentric videos, trained solely on large-scale unlabeled data. By aligning video depth with 4D constraints, it achieves promising zero-shot results in dense, generalizable scene reconstruction. Additional discussion of limitations and future directions of EgoMono4D could be found in Appendix~\ref{app: limitation_future}.

\section{Acknowledgment} 

This work is supported by the National Key R\&D Program of China (2022ZD0161700), National Natural Science Foundation of China (62176135), Shanghai Qi Zhi Institute Innovation Program SQZ202306 and the Tsinghua University Dushi Program, the grant of National Natural Science Foundation of China (NSFC) 12201341.

{
    \small
    \bibliographystyle{ieeenat_fullname}
    \bibliography{main}
}

\newpage
\
\newpage

\section*{Appendix}

\appendix

\begin{table*}[!t]
    \centering
    \begin{tabular}{lcccccc}
    \toprule
    \multicolumn{7}{c}{\textbf{Training}} \\ 
    \midrule
    \textbf{Datasets}     & \makecell{\textbf{\# of frames}} & \makecell{\textbf{Data Split}}  & \makecell{\textbf{Camera Motion}} & \textbf{Action}  & \textbf{Scene}   & \textbf{Note} \\ 
    \midrule
    H2O          & 40K          & Original Split    & Small         & Simple  & Clean   & Bi-manual \\ 
    HOI4D        & 540K         & Room ID           & Medium        & Simple  & Clean   &  \\ 
    FPHA         & 73K          & Task              & Medium        & Complex & Clutter &  \\ 
    EgoPAT3D     & 823K         & Scene ID          & Large         & Medium  & Medium  & Only pick \& place \\ 
    Epic-Kitchen & 9.7M         & Scene ID       & Large         & Complex & Clutter &  \\ 
    \midrule
    \multicolumn{7}{c}{\textbf{Evaluation}} \\ 
    \midrule
    \textbf{Datasets}     & \makecell{\textbf{\# of frames}} & \textbf{Label}        & \makecell{\textbf{Camera Motion}} & \textbf{Action}  & \textbf{Scene}   & \textbf{Note} \\ 
    \midrule
    H2O          & 8K          & Calibration       & Small         & Simple  & Clean   & Bi-manual \\ 
    HOI4D        & 12K         & SfM            & Medium        & Simple  & Clean   & Contain noise \\ 
    ARCTIC-HOI   & 13K          & Mocap             & Medium        & Complex & Clean   & \makecell{Only hand and object label} \\ 
    POV-Surgery  & 26K         & Synthesis         & Large         & Medium  & Clutter & \makecell{Unrealistic texture} \\ 
    \bottomrule
    \end{tabular}
    \caption{Comparison of Different Datasets for Training and Evaluation}
    \label{tab:dataset_comparison}
\end{table*}


\noindent This appendix file provides:

\begin{itemize}
    \item \ref{app: datasets}. Details of Datasets 
    \item \ref{app: unidepth}. Demonstration of UniDepth Backbone
    \item \ref{app: metric}. Details of Evaluation Metrics
    \item \ref{app: flow_result}. Results of Long-term 3D Scene Flow Recovery
    \item \ref{app: inference_speed}. Results of Model Inference Speed. 
    \item \ref{app: infer_ablation}. Ablation of Inference Strategy
    \item \ref{app: depth_camera}. Depth and Camera Results
    \item \ref{app: more_vis}. More Reconstruction Results Visualization
    \item \ref{app: qualitative_comp}. Qualitative Comparison with Baseline
    \item \ref{app: limitation_future}. Limitations and Future Directions
\end{itemize}

For video visualization and online interactive demonstration, please refer to \href{https://egomono4d.github.io/}{https://egomono4d.github.io/}.

\section{Details of Datasets} \label{app: datasets}

Figure 3 in the main paper visualizes samples from various datasets. Detailed information about the datasets is shown in Table~\ref{tab:dataset_comparison}. The data encompass different types of egocentric videos, with variations in camera motion (small/large), action complexity (simple/complex), and scene conditions (clean/cluttered).

For training, we use a combination of H2O\cite{h2o} (40K frames), HOI4D\cite{hoi4d} (540K frames), FPHA\cite{fpha} (73K frames), EgoPAT3D\cite{egopat3d} (823K frames), and Epic-Kitchen\cite{epic_kitchen} (9.7M frames). The final training dataset contains a total of 11.2M frames, with Epic-Kitchen dominating the data (about 85\%), providing large-scale diversity and a wide range of behavior modes. The other datasets contribute unique scene characteristics or behavior patterns. For instance, H2O\cite{h2o} exclusively contains bi-manual operations with small camera motion in clean table scenes. Approximately 5\% of the data is allocated for model validation.

For evaluation, we note a scarcity of egocentric datasets offering both high-quality depth and precise camera labels. We strongly encourage the computer vision and HOI communities to collect or synthesize larger-scale RGBD datasets with accurate pose annotations. We selected four datasets that meet the necessary criteria: H2O\cite{h2o}, HOI4D\cite{hoi4d}, ARCTIC\cite{arctic}, and POV-Surgery\cite{pov_surgery}. The sources for camera labels are provided in Table~\ref{tab:dataset_comparison}. 

H2O\cite{h2o} and HOI4D\cite{hoi4d} feature simple scenes and actions and are used as test benchmarks for in-domain prediction performance. For zero-shot generalization evaluation, we use ARCTIC\cite{arctic} and POV-Surgery\cite{pov_surgery}. ARCTIC\cite{arctic} only provides labels for hands and objects, which is why we refer to it as ARCTIC-HOI, It is primarily used to assess the reconstruction quality of HOI. Both ARCTIC-HOI and POV-Surgery present significant challenges for EgoMono4D, as they exhibit a large domain gap from the training data. (1) For ARCTIC-HOI, the hand and object components occupy a much larger portion of the images compared to the training data. (2) For POV-Surgery, the dataset's unrealistic textures create a substantial visual domain gap, and the surgical scenes were not encountered during training.

\section{Demonstration of UniDepth Backbone} \label{app: unidepth}

Our architecture builds upon the UniDepth backbone \cite{unidepth}, an encoder-decoder architecture depth estimator. UniDepth decouples the tasks of depth and camera estimation by transforming the scene from Cartesian coordinates to a pseudo-spherical representation, which enables dense camera prediction in spherical space. It then incorporates scene scale information from the camera prediction into the depth estimation module using Laplace Spherical Harmonic Encoding (SHE) and a cross-attention mechanism \cite{attention, vit}. For more details, please check out the original paper \citet{unidepth}.

\section{Details of Evaluation Metrics} \label{app: metric}

We provide the mathematical definitions of the metrics used to evaluate the performance of 3D point cloud sequence reconstruction and long-term 3D scene flow recovery. 

\subsection{Metrics for Pointclouds Sequence} 

We follow \citet{pcd_metric} in using 3D Chamfer Distance (CD, measured in millimeters) and the 3D Pointclouds F-score (F, measured as a percentage \%) to evaluate shape similarity. For implementation, we leverage the Kaolin library \cite{kaolin}. Given the ambiguity in the scale of pointclouds, we first align the predicted point cloud sequence to the ground truth using an estimated best-aligned global scaled SE(3) transformation $(s,R,T)$.

\vspace{0.25cm}

\noindent \textbf{3D Chamfer Distance (CD, mm).}\ \ Given the predicted per-frame pointclouds $P \in R^{N\times 3}$ and the ground-truth $G\in R^{N\times 3}$, where $N$ is the number of points, the 3D Chamfer Distance (CD) is defined as:
\begin{equation}
    CD(R,G) = \sum_{x\in G} \mathop{\min}\limits_{y\in R} ||x-y|| + \sum_{y\in R}  \mathop{\min}\limits_{x\in G} ||x-y||
\end{equation}

\vspace{0.25cm}

\noindent \textbf{3D Pointclouds F-score (F).}\ \ The 3D F-score combines precision and recall to provide a balanced evaluation of the predicted surface quality. In the context of 3D pointclouds, precision measures how many points from the predicted surface are close to the ground truth surface, while recall measures how many ground truth points are captured by the predicted surface. Given a distance $\delta$ as the positive threshold, the precision $P_{\delta}(R,G)$, recall $R_{\delta}(R,G)$ and F-score $F_{\delta}(R,G)$ are defined as:
\begin{align}
&P_{\delta}(R,G) = \frac{1}{|R|} \sum_{y\in R} [d_{y\to G} < \delta] \\
&R_{\delta}(R,G) = \frac{1}{|G|} \sum_{x\in G} [d_{x\to R} < \delta] \\ 
& F_{\delta}(R,G) = \frac{2P_{\delta}(R,G)R_{\delta}(R,G)}{P_{\delta}(R,G) + R_{\delta}(R,G)}
\end{align}

\begin{table*}[!t]
\centering
\small
\setlength{\tabcolsep}{3pt} 
\newcolumntype{C}{>{\centering\arraybackslash}X} 
\begin{tabularx}{\textwidth}{lCCCCCCCCCCCCCCCC}
\toprule
                      & \multicolumn{4}{c}{\textbf{HOI4D}}    & \multicolumn{4}{c}{\textbf{H2O}}  & \multicolumn{4}{c}{\textbf{POV-Surgery$^{\dag
}$}}   & \multicolumn{4}{c}{\textbf{ARCTIC-HOI$^{\dag
}$}}    \\ 
                      \cmidrule(lr){2-5} \cmidrule(lr){6-9} \cmidrule(lr){10-13} \cmidrule(lr){14-17}
                      & {ADE$\downarrow$} & {FDE$\downarrow$} & {P$_{5}\uparrow$} & {P$_{10}\uparrow$} & {ADE$\downarrow$} & {FDE$\downarrow$} & {P$_{5} \uparrow$} & {P$_{10}\uparrow$} &   {ADE$\downarrow$} & {FDE$\downarrow$} & {P$_{5} \uparrow$} & {P$_{10}\uparrow$} &  {ADE$\downarrow$} & {FDE$\downarrow$} & {P$_{5}\uparrow$} & {P$_{10} \uparrow$} \\
\midrule
Modularized Version & 88.8	&94.8 & 23.1	&69  & \underline{40.9} &	\underline{42.8} & \underline{75.8} &	\textbf{97.0} & 504.5	&767.6&  3.2 & 13.6  & 102.0 &	154.6&  33.8&	62.0 \\
DS+UniDepth~\cite{droid_slam} & \underline{64.0}  & 75.1  & \underline{54.0} & 82.5 & 46.4 & 48.6 & 67.6 & 95.0 & 168.9 & 199.7 & 13.2 & 39.8 & \underline{53.9} & \textbf{72.1} & \textbf{60.1} & \underline{86.8}  \\
DUSt3R~\cite{dust3r} & 79.2	& 75.8	&	47.7	&75.1	&71.7	&71.5	&54.3	&75.9 & 412.8	 & 407.2	& \underline{14.8} &	\underline{44.9}	&214.3	&181.6 &	52.2 &	82.3	 \\
MonSt3R~\cite{monst3r} & 70.2	& 71.0 & 50.7	& 80.6 &	96.3 &	97.1 & 27.6	& 67.3 &	201.8 &	208.9 &	8.4 & 31.8 &	83.9	&106.3	&	28.4 &	69.9  \\
Align3R~\cite{align3r} & 66.3	& \underline{67.8} & 53.4	& \underline{83.7} &	75.3 &	75.5 & 39.2	& 79.7  &	\underline{157.6} &	\underline{163.7} &	14.5 & 43.1 &	64.8	& 83.3	&	43.8 &	83.5  \\
CUT3R~\cite{cut3r}      & 72.5 & 72.6 & 49.8 & 81.3 & 50.7 & 57.4 & 67.9 & 91.3 & 304.6 & 353.0 & 4.6 & 19.3  & 84.3 & 106.2 & 27.4 & 69.9    \\
\midrule
EgoMono4D (Ours)  & \textbf{57.3}   & \textbf{60.6}  & \textbf{59.2} &\textbf{87.0} & \textbf{37.0} & \textbf{38.9} &\textbf{77.0} & \underline{96.4} & \textbf{125.0} & \textbf{138.9} & \textbf{19.1} & \textbf{55.2 } & \textbf{53.8} & \textbf{72.1} & \underline{57.2} & \textbf{87.3} \\
\bottomrule
\end{tabularx}
\caption{The evaluation results for long-term 3D scene flow recovery are presented, with ADE (mm), FDE (mm), and Precision ($P_{\delta}$, \%). $\dag$ denotes zero-shot generalization for EgoMono4D. For ARCTIC-HOI, the evaluation focuses solely on hand-object recovery quality. Overall, EgoMono4D significantly outperforms the other baselines.}
\label{tab: flow_reconstruction}
\end{table*}

\begin{table}[!t]
\centering
\footnotesize 
\setlength{\tabcolsep}{4pt} 
\begin{tabular}{lcccc}
\toprule
(Only for HOI)           & {ADE$\downarrow$} &  {FDE$\downarrow$} & {P$_5$$\uparrow$} & {P$_{10}$$\uparrow$} \\ \midrule
HOI4D      & 79.3 / \textbf{76.6} & 84.6 / \textbf{78.4} & \textbf{48.3} / 46.2 & 77.8 / \textbf{81.0}    \\
H2O         & 43.6 / \textbf{40.5} & 45.8 / \textbf{41.4} & 68.6 / \textbf{71.2} & 95.7 / \textbf{98.2}    \\
POV-Surgery$^{\dag}$ & 196.0 / \textbf{192.2} & 213.4 / \textbf{206.4} & \textbf{9.9} / \textbf{9.9}   & 32.4 / \textbf{32.8}   \\ \bottomrule
\end{tabular}
\caption{\footnotesize Additional long-term 3D scene flow results on the HOI part. Values are presented \textbf{in the format 'DS+UniDepth / EgoMono4D (Ours)'}. Both models demonstrate comparable performance, with EgoMono4D showing a modest advantage.} 
\label{tab: pcd_hoi_comparison}
\end{table}

\subsection{Metrics for Long-term 3D Scene Flow}

Long-term 3D scene flow refers to predicting the future trajectories of multiple 3D query points in the pointclouds of the first frame \cite{general_flow}. Following previous works \cite{tapvid3d, general_flow, 3d_hand_traj}, we evaluate the precision of 3D flow recovery using three metrics: Average Displacement Error (ADE, measured in millimeters), Final Displacement Error (FDE, measured in millimeters), and Precision under Distance (P, measured as a percentage \%). The 3D flow is generated by interpolating between the predicted and ground-truth pointclouds based on 2D tracking from CoTracker \cite{cotracker}. Before evaluation, we filter out trajectories affected by noise from flying pixels \cite{fly_pixel} using ground-truth depth information. To align the scale of scenes and the initial position of the 3D query points, we first perform a best-aligned scaled SE(3) transformation between the ground-truth and predicted pointclouds for the first frame.

\vspace{0.25cm}

\noindent \textbf{Average Displacement Error (ADE, mm).}\ \ Given the predicted 3D flow $F \in R^{T \times N \times 3}$ and ground-truth flow $G \in R^{T \times N \times 3}$, where $T$ represents the number of frames and $N$ represents the number of trajectories, ADE measures the average displacement across all timestamps.

\begin{equation}
    ADE(F,G) = \frac{1}{N} \sum_{i=1}^N ||F_i - G_i||
\end{equation}

\vspace{0.25cm}

\noindent \textbf{Final Displacement Error (FDE, mm).}\ \ FDE measures the displacement of the final timestamp.
\begin{equation}
    FDE(F,G) = ||F_{T-1} - G_{T-1}||
\end{equation}

\vspace{0.25cm}

\noindent \textbf{Precision under Distance (P, \%).}\ \ The precision metric measures the average percentage of points with an error within $\delta$ centimeters (cm).
\begin{equation}
    P_{\delta} = \frac{1}{N} \sum_{i=1}^N [||F_i - G_i|| < \delta]
\end{equation}

\section{Long-term 3D Scene Flow Recovery Results}  \label{app: flow_result}

\begin{figure*}[!t]
  \centering
   \includegraphics[width=1.0\linewidth]{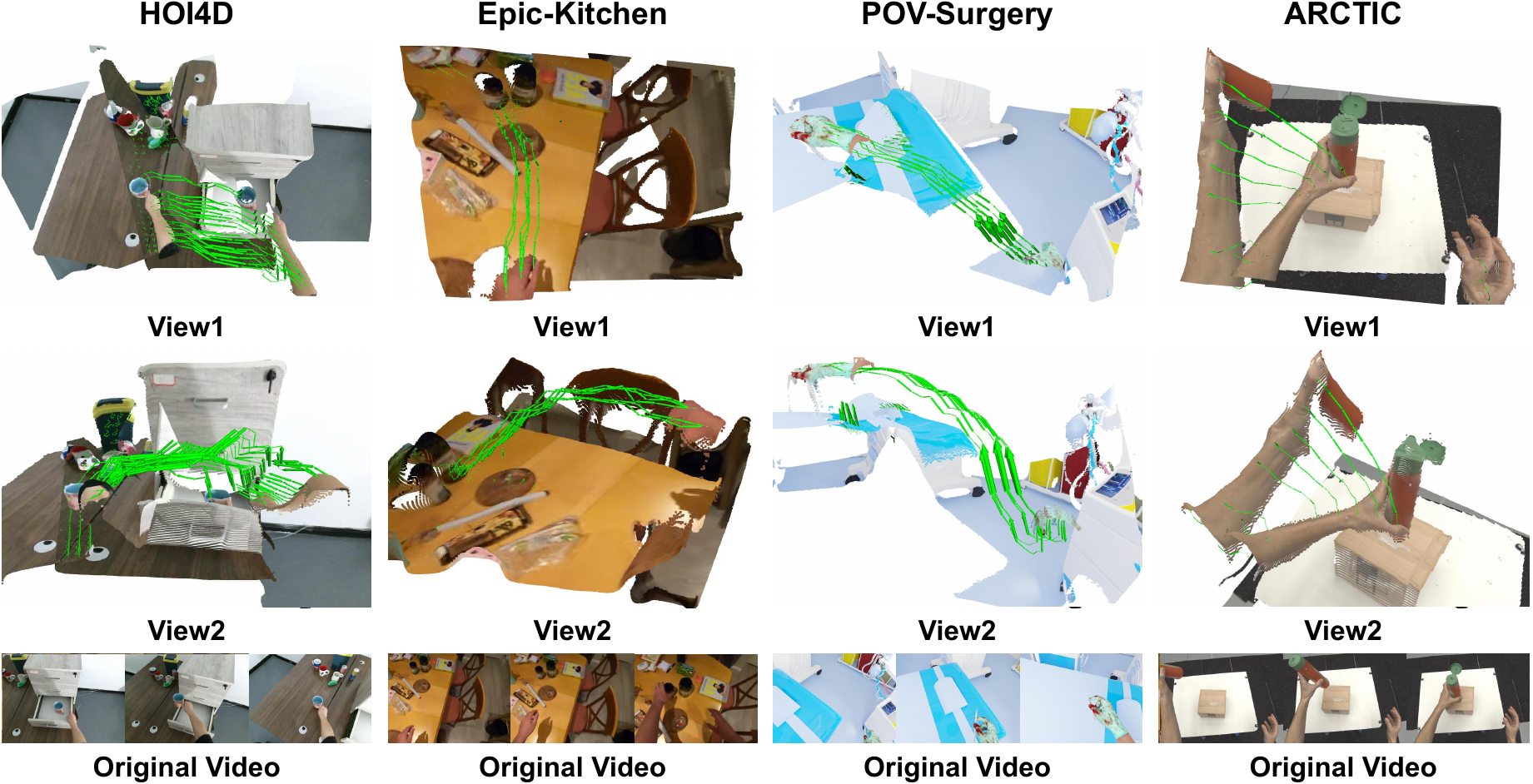}
   \caption{The visualization of the long-term 3D scene flow recovery. For clarity, we display only the pointclouds from the first and last frame. The green arrows representing the estimated 3D flow. EgoMono4D successfully recovers the motion of dynamic parts while maintaining other regions static to some extent.}
   \label{fig:4_flow_visualization}
\end{figure*}

\begin{table*}[!t]
\centering
\renewcommand{\arraystretch}{1.0} 
\setlength{\tabcolsep}{3pt} 
\begin{tabular}{>{\centering\arraybackslash}p{1.5cm}|>{\centering\arraybackslash}p{1.5cm}|>{\centering\arraybackslash}p{1.5cm}>{\centering\arraybackslash}p{1.5cm}>{\centering\arraybackslash}p{1.5cm}>{\centering\arraybackslash}p{1.5cm}|>{\centering\arraybackslash}p{1.5cm}>{\centering\arraybackslash}p{1.5cm}>{\centering\arraybackslash}p{1.5cm}>{\centering\arraybackslash}p{1.5cm}}
\toprule 
\multirow{2}{*}{\textbf{$N_w$}} & \multirow{2}{*}{\textbf{$N_o$}} & \multicolumn{4}{c|}{\textbf{HOI4D}} & \multicolumn{4}{c}{\textbf{POV-Surgery}} \\ \cline{3-10} 
                             &                              & \textbf{CD$\downarrow$} \rule{0pt}{12pt} & \textbf{F$_{1}\uparrow$} & \textbf{F$_{2.5}\uparrow$} & \textbf{F$_{5}\uparrow$}  & \textbf{CD$\downarrow$}  & \textbf{F$_{1}\uparrow$} & \textbf{F$_{2.5}\uparrow$} & \textbf{F$_{5}\uparrow$}  \\ \midrule 
\textbf{4}                   & \textbf{1}                   & 5.9       & 27.9      & 59.6            & 83.1                  & 33.8        & 13.5      & 32              & 53.9                  \\ \midrule
\textbf{2}                   & \textbf{1}                   & 17.9      & 11.5       & 29.4            & 51.6                 & /             & /               & /              & /              \\ 
\textbf{8}                   & \textbf{1}                   & 6.4         & 26.0      & 55.6            & 80.0                 & 35.2      & 12.6      & 30.8            & 53.1                    \\ 
\textbf{12}                  & \textbf{1}                   & 6.7       & 25.1     & 53.5            & 78.4                    & /             & /               & /              & /              \\ \midrule
\textbf{4}                   & \textbf{2}                   & 5.9     &  27.9       & 59.7            & 83.1                   & 34.6     & 13.4      & 32              & 53.9                    \\ 
\textbf{4}                   & \textbf{3}                   & 5.9     & 27.9        & 59.7            & 83.1                   & 35.0        & 13.4    & 31.9            & 53.8                    \\ \midrule
\textbf{8}                   & \textbf{4}                   & /            & /               & /              & /              & 36.3     & 16.6       & 30.5            & 52.6                    \\ \bottomrule 
\end{tabular}
\caption{Comparison of pointclouds sequence reconstruction results across different window sizes ($N_w$) and overlapping sizes ($N_o$). Our model demonstrates robustness to variations in $N_o$, while maintaining consistency in $N_w$ between training and inference is essential.}
\label{tab:nonw_result}
\end{table*}

\begin{table}[!t]
\centering
\small
\begin{tabular}{c|>{\centering\arraybackslash}p{1.25cm}>{\centering\arraybackslash}p{1.25cm}>{\centering\arraybackslash}p{1.25cm}}
\toprule
\multicolumn{1}{l}{\multirow{2}{*}{}} & \multicolumn{3}{c}{\textbf{POV-Surgery (Video Depth)}} \\ 
\cmidrule(lr){2-4} 
                                       & AbsRel$\downarrow$ & \textbf{$\delta_{0.05}\uparrow$} & \textbf{$\delta_{0.1}\uparrow$}   \\ 
\midrule
UniDepth~\cite{unidepth}           & \textbf{11.9}             & 40.8           & \textbf{64.7}            \\ 
DUSt3R~\cite{dust3r}                                & 19.7                & 31.5           & 51.9           \\ 
MonSt3R~\cite{monst3r}                                & 18.6             & 33.6           & 52.8            \\  
Align3R~\cite{align3r}                                & 13.3             & \textbf{41.7}           & 62.6            \\ \midrule
EgoMono4D (Ours)                               & \underline{12.6}                 & \underline{41.1}           & \underline{63.9}            \\ 
\bottomrule
\end{tabular}

\begin{tabular}{c|>{\centering\arraybackslash}p{1.2cm}>{\centering\arraybackslash}p{1.2cm}>{\centering\arraybackslash}p{1.2cm}}
\multicolumn{1}{l}{\multirow{2}{*}{}} & \multicolumn{3}{c}{\textbf{POV-Surgery (Camera Poses) \rule{0pt}{12pt}}} \\ 
\cmidrule(lr){2-4} 
                                       & ATE$\downarrow$ & RPE-T$\downarrow$ & RPE-R$\downarrow$   \\ 
\midrule
Modularized Version                                & 47.03       &  4.10           & 4.18                  \\ 
DS+UniDepth~\cite{droid_slam}           & 9.05        &   4.17        & 0.39                     \\ 
MonSt3R~\cite{monst3r}                                & \underline{6.63}       & \underline{2.41}           & \underline{0.26}                    \\ 
Align3R~\cite{align3r}                                & \textbf{6.35}       & \textbf{2.34}           & \textbf{0.23}                    \\
CUT3R~\cite{cut3r}                                & 8.11       & 2.92           & 0.31 \\
\midrule
EgoMono4D (Ours)                               &  11.54       & 4.01           &  0.43                \\ 
\bottomrule
\end{tabular}

\caption{Result of POV-Surgery video depth and camera poses estimation. It shows that our model does not outperform others in estimating these geometric variables. Instead, our model focus on improving their consistency in 3D space, which leads to better pointclouds sequence reconstruction.}
\label{tab: depth_camera_result}
\end{table}

\begin{figure*}[!t]
  \centering
   \includegraphics[width=1.0\linewidth]{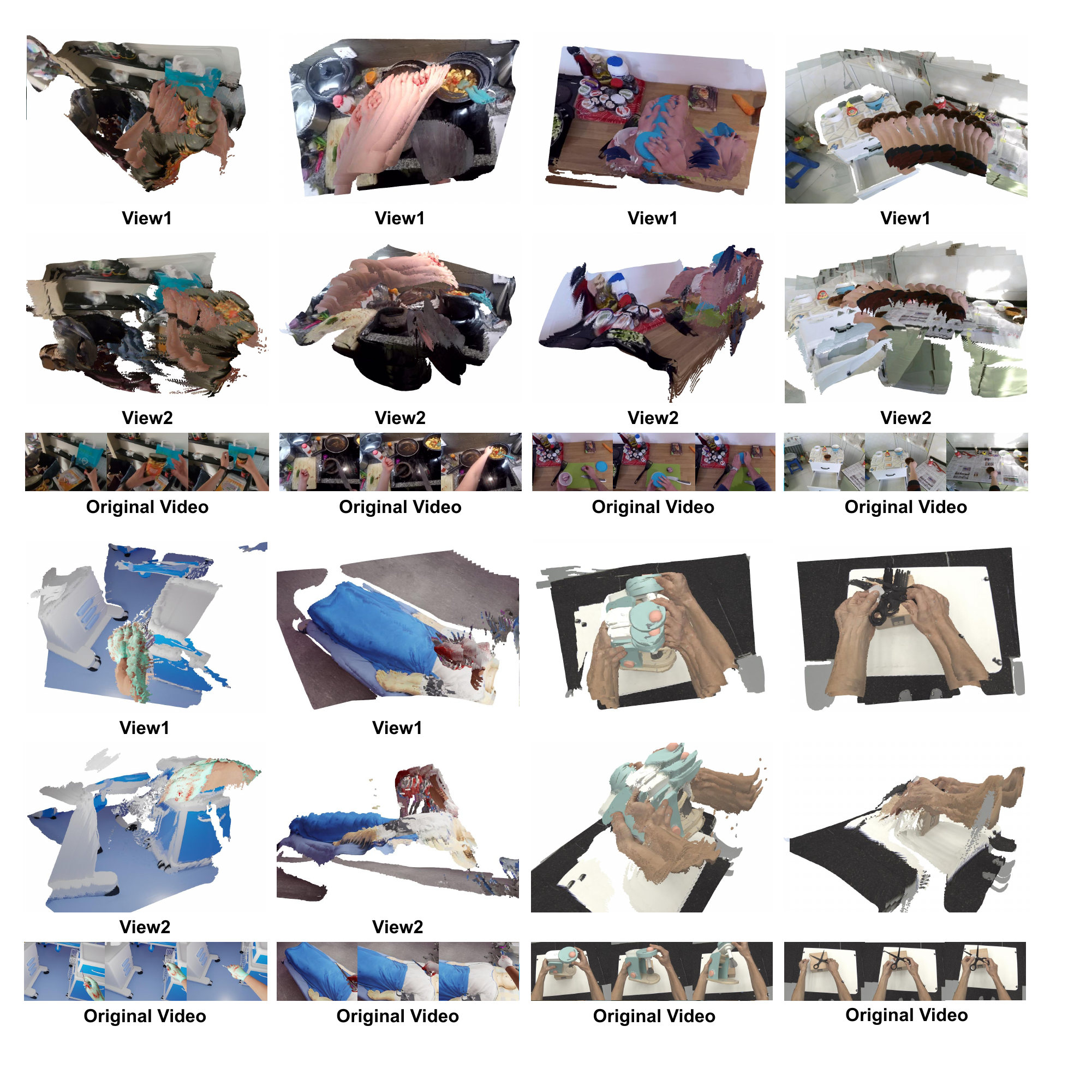}
   \caption{More visualization of pointclouds sequence reconstruction results from EgoMono4D.}
   \label{fig:2s_vis}
\end{figure*}

\begin{figure*}[!t]
  \centering
   \includegraphics[width=0.95\linewidth]{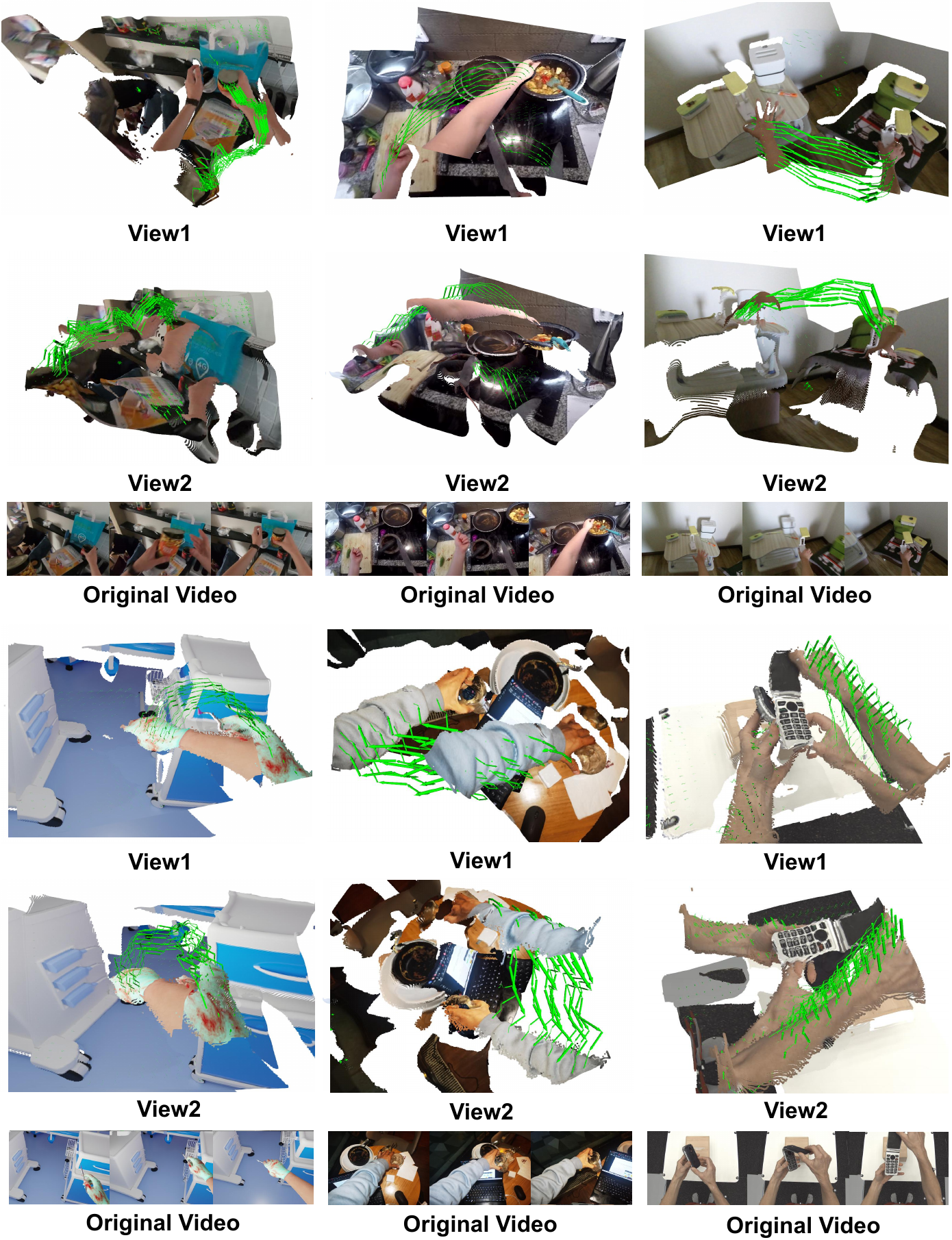}
   \caption{More visualization of long-term 3D scene flow recovery results from EgoMono4D.}
   \label{fig:2s_vis_flow}
\end{figure*}

\begin{figure*}[!t]
  \centering
   \includegraphics[width=0.95\linewidth]{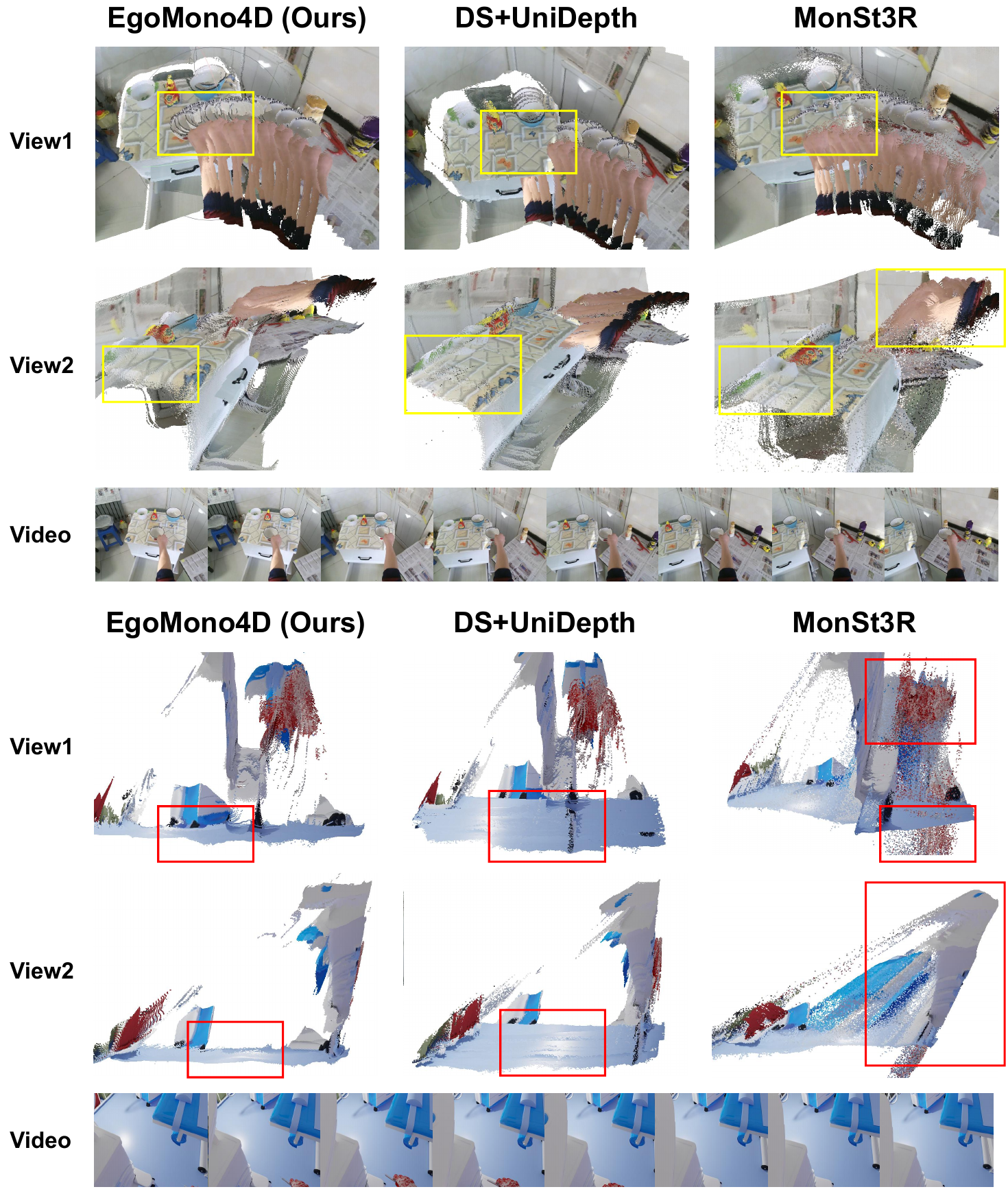}
   \caption{Qualitative comparison between EgoMono4D and other baseline methods. EgoMono4D demonstrates superior performance compared to baseline methods in both static scene reconstruction and dynamic HOI motion recovery. DS+UniDepth \cite{droid_slam, unidepth} struggles with static scene alignment, while MonSt3R \cite{monst3r} exhibits limitations in accurately reconstructing hand geometry.}
   \label{fig:2s_vis_quality}
\end{figure*}

\begin{figure*}[!t]
  \centering
   \includegraphics[width=0.95\linewidth]{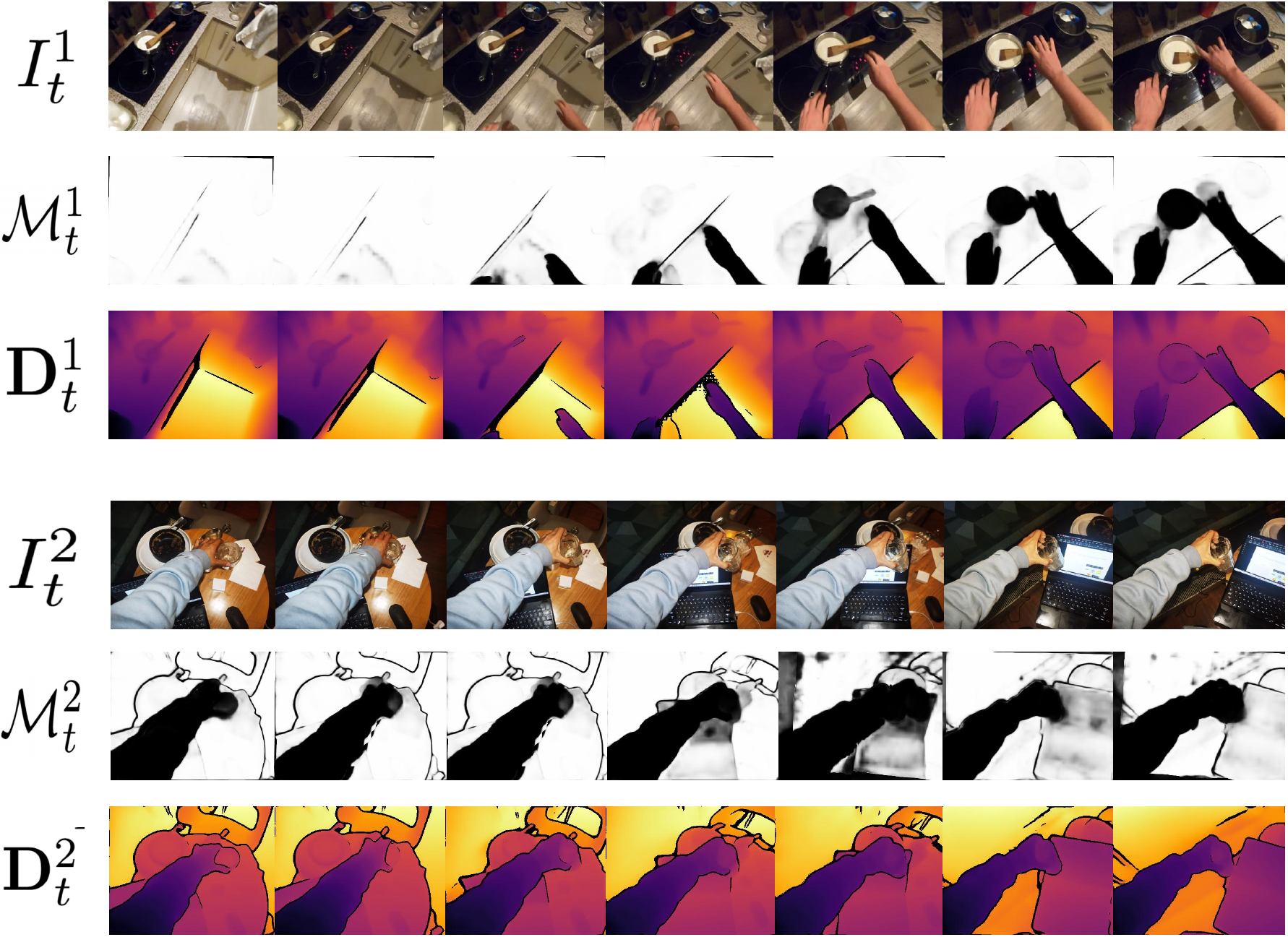}
   \caption{Visualization of original video $I_t$ and the predicted procrustes-alignment confidence maps $\mathcal{M}_t$ and video depth $D_t$ from EgoMono4D.}
   \label{fig:2s_vis_geo}
\end{figure*}

\noindent \textbf{Task}\ \ Long-term 3D scene flow \cite{3d_scene_flow} captures both the structure and dynamics of egocentric scenes in a compact format, making it valuable for various applications such as perception \cite{scene_flow_segmentation}, autonomous driving \cite{scene_flow_autodrive}, and robot learning \cite{general_flow, flowbot3d, toolflownet}. Given a video and a set of query points in the first frame, the 3D flow \cite{general_flow, tapvid3d} represents the future trajectory of each query point in 3D space. Since egocentric datasets lack explicit 3D flow labels, we first employ CoTracker \cite{cotracker}, a high-precision pixel tracker, to generate 2D long-term tracking (with a 35$\times$35 grid of query points). These 2D trajectories are then unprojected using ground-truth pointclouds sequence to create 3D trajectory labels. We use the same method to obtain predictions for models. 

\vspace{0.25cm}

\noindent \textbf{Metric}\ \ To align the scale of scenes
, we first perform a best-aligned scaled SE(3) transformation between the ground-truth and predicted pointclouds for the first frame. We adopt metrics from related works \cite{general_flow, tapvid3d, 3d_hand_traj, shape-of-motion}, including Average Displacement Error (ADE, mm), Final Displacement Error (FDE, mm), and Precision under Distance (P, $\%$). We use the notation $P_{\delta}$ to represent precision with a $\delta$ centimeter (cm) threshold. Details are demonstrated in Appendix~\ref{app: metric}.

\noindent \textbf{Result}\ \ Table~\ref{tab: flow_reconstruction} shows that our model outperforms all baselines for long-term 3D scene flow recovery on average. The visualizations are shown in Figure~\ref{fig:4_flow_visualization}. DS+UniDepth performs well in hand-object motion estimation on the ARCTIC-HOI dataset. We also compared the HOI estimation of both models on three other datasets, with results in Table~\ref{tab: pcd_hoi_comparison}. Both models perform similarly, with EgoMono4D shows a modest advantage. However, DS+UniDepth's performance drops notably in full-scene estimation (e.g., POV-Surgery dataset) due to inconsistencies and accumulated errors between modules. Other models perform worse overall. Figure~\ref{fig:4_flow_visualization} illustrates the estimated long-term 3D scene flow. It can be seen that EgoMono4D successfully reconstructs 3D dynamics across diverse scenes. 

\section{Inference Speed Results} \label{app: inference_speed}

\begin{figure}[t]
  \centering
   \includegraphics[width=0.85\linewidth]{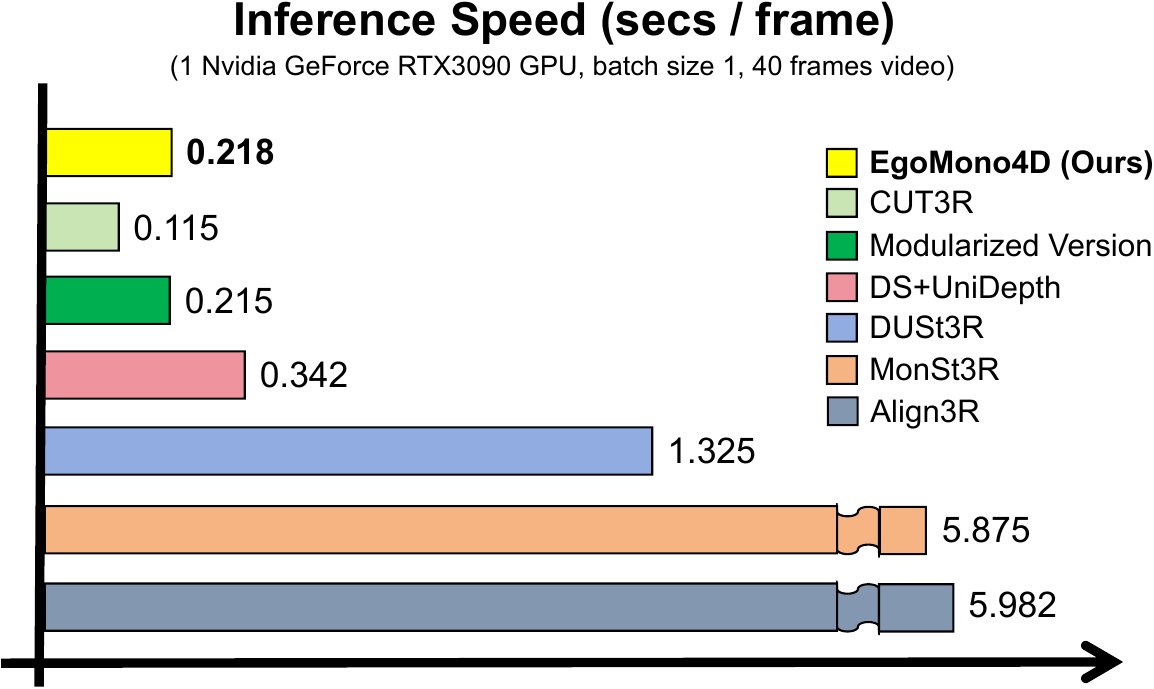}
   \caption{Inference speed comparison of different methods.}
   \label{fig:6-speed}
\end{figure}

We evaluate the inference speed of different models (for 40 frames video). All measurements are conducted on a single NVIDIA GeForce 3090 GPU with a batch size of 1. Results are shown in Figure~\ref{fig:6-speed}. Our model achieves 0.218 secs/frame speed, which is only slower than its modularized version~\cite{mapfree_vr} (0.215 secs/frame) and CUT3R~\cite{cut3r} (0.115 sec/frame). However, the speed difference is very small, while our model achieves better reconstruction quality with large advantage.

\section{Ablation on Inference Strategy}\label{inference_strategy}  \label{app: infer_ablation}

During inference, our model processes $N_w$ frames in a single feed-forward prediction. Theoretically, the window size $N_w$ can be any value greater than 1. For videos with more frames than $N_w$, the overlapping size $N_o$ between neighboring windows must also be determined. By default, we set $N_w = 4$ (consistent with training) and $N_o = 1$ (to optimize inference speed). We evaluate the impact of $N_w$ and $N_o$ on reconstruction performance using the pointclouds sequence reconstruction task on HOI4D \cite{hoi4d} and POV-Surgery \cite{pov_surgery}, with results shown in Table~\ref{tab:nonw_result}. 

For the window size $N_w$, maintaining consistency between training and inference is crucial, likely because the video adapter is trained specifically to fuse 4 frames. Regarding overlapping size $N_o$, the model exhibits comparable performance for $N_o = 1, 2, 3$. Therefore, we select $N_o = 1$ to maximize inference speed.

\section{Depth and Camera Results}  \label{app: depth_camera}

We further evaluate the video depth and camera poses estimation performance on POV-Surgery \cite{pov_surgery}, using the metrics from MonSt3R \cite{monst3r}. The results are presented in Table~\ref{tab: depth_camera_result}. Our model does not outperform other methods in estimating these independent geometric variables. The depth and camera prediction performance of EgoMono4D is only comparable with other baseline methods. Instead, it enhances the consistency of them in 3D space, leading to improved pointclouds sequence reconstruction results. UniDepth \cite{unidepth} achieves the best depth estimation, while Align3R \cite{align3r} provides the most accurate camera poses.

We believe this is due to two main reasons: (1) Our method is primarily optimized for 4D pointclouds reconstruction, rather than single geometry variable estimation. While it may not achieve the highest accuracy for individual variables, it ensures greater consistency and alignment among them. (2) Compared with supervised methods, self-supervised approach does not show a clear advantage when focusing solely on single-variable estimation, as indicated by previous work~\cite{kick}. This is because there are more datasets available for single-variable estimation. We believe that using more advanced pretrained models and training on 0larger datasets may address this in the future~\cite{kick, kick++}.

\section{More Reconstruction Results Visualization} \label{app: more_vis}

We provide more visualization of pointclouds sequence reconstruction results of EgoMono4D in Figure~\ref{fig:2s_vis}. More visualization of recovered long-term 3D scene flows are shown in Figure~\ref{fig:2s_vis_flow}. Finally, we provide the visualization of intermediate geometric variables in Figure~\ref{fig:2s_vis_geo}.

\section{Qualitative Comparison with Baseline}  \label{app: qualitative_comp}

We qualitatively compare EgoMono4D with other baseline methods on the dense point cloud sequence reconstruction task. Visual comparisons are presented in Figure~\ref{fig:2s_vis_quality}. EgoMono4D demonstrates superior performance compared to baseline methods in both static scene reconstruction and dynamic HOI motion recovery. DS+UniDepth \cite{droid_slam, unidepth} struggles with static scene alignment, while MonSt3R \cite{monst3r} exhibits limitations in accurately reconstructing hand geometry and dynamics.

\begin{figure}[!t]
  \centering
   \includegraphics[width=1.0\linewidth]{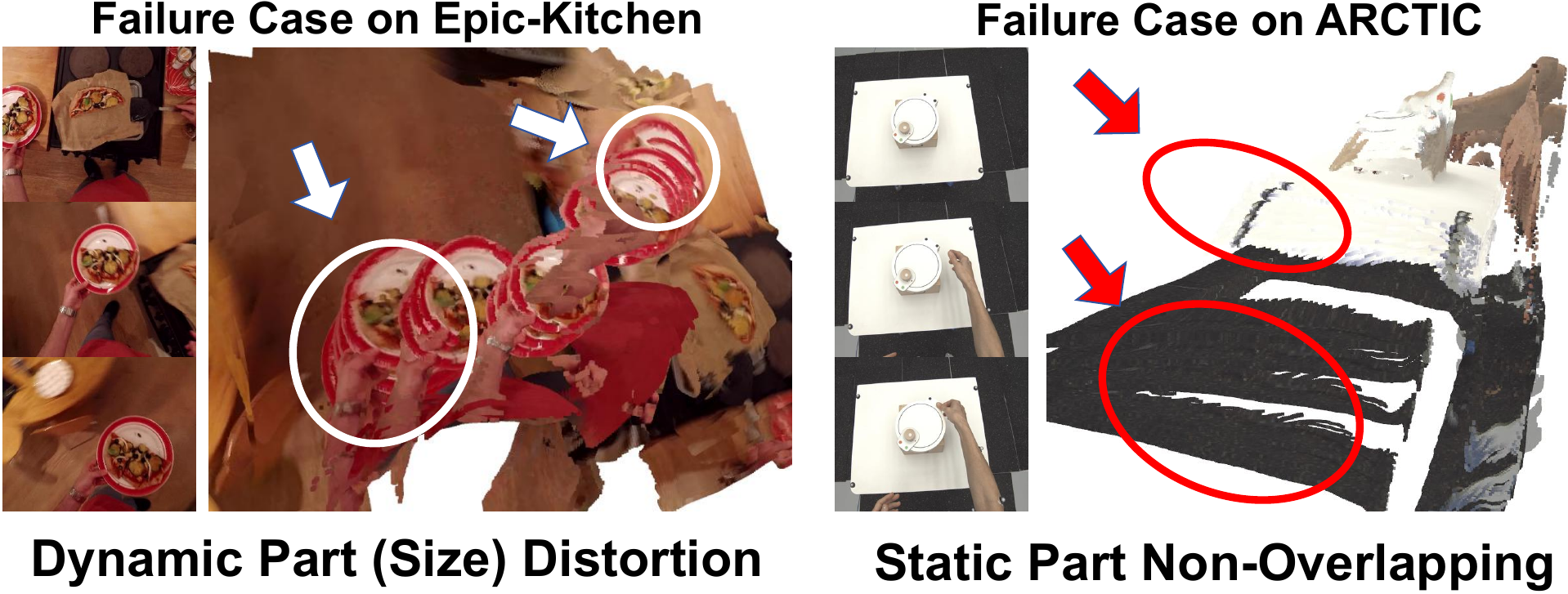}
   \caption{Two typical failure cases of EgoMono4D arise from inherent inconsistencies in UniDepth shape supervision.}
   \label{fig:5_fail}
\end{figure}

\section{Limitations and Future Directions}  \label{app: limitation_future}

While EgoMono4D achieves impressive results in fast, dense, and generalizable dynamic HOI scene reconstruction, it still faces challenges with shape misalignment (see Figure 5 in the main paper). This issue arises from inherent inconsistencies in UniDepth’s \cite{unidepth} shape regularization and can be divided into two key problems. (1) Dynamic part (size) distortion: since we only regularize the shape of the dynamic part based on per-frame pointclouds predictions from UniDepth, the relative size of the dynamic part compared to the static part is determined by UniDepth's predictions. Any inaccuracy in this relative size may be carried over to EgoMono4D. (2) Static part misalignment: if the original predicted shapes of the static areas from UniDepth differ significantly between frames, it becomes difficult to adjust and align them consistently.

Although precise posed datasets are limited, there are more datasets available with depth labels. Training on these datasets with ground-truth shape supervision, or using a combination of labeled and unlabeled datasets, could help address these issues. Improvements in monocular depth estimation and intrinsic parameter estimation may also alleviate these problems.

Another limitation of EgoMono4D is that it currently supports video reconstruction only for videos with an FPS above a certain threshold. This is due to its reliance on the off-the-shelf optical flow estimation module \cite{gmflow}, which may perform poorly with sparse views. Integrating correspondence, matching, or optical flow prediction directly into the model to enable fully end-to-end training could address this limitation \cite{vggsfm, mast3r}. Sparse-view data also needs to be incorporated during training \cite{dust3r} to solve this problem.

Additionally, our model currently only supports a resolution of  288$\times$384. Training models at higher resolutions could enable more flexible applications. Our model also fails when the dynamic portion of the image is too large or contains too many moving objects beyond the ones being manipulated. Although we focus on egocentric HOI scenes, our training paradigm could be extended to more general cases. We plan to train this general scene model using motion mask priors derived from motion segmentation \cite{tmo_motion, flowsam}, salient video segmentation \cite{ufo}, and semantic segmentation \cite{groundingdino, sam}. We also encourage the community to propose more synthetic datasets \cite{pointodyssey, adt} to explore supervised learning approaches \cite{dust3r, monst3r}.

\end{document}